\documentclass{article} 
\usepackage{iclr2026_conference,times}

\usepackage{booktabs}   
\usepackage{multirow}   
\usepackage{graphicx}   
\usepackage{pifont}     
\usepackage{caption}    
\usepackage{adjustbox}
\usepackage{enumitem}

\usepackage{makecell}      
\usepackage[table]{xcolor}   

\usepackage{graphicx} 
\usepackage{tikz}     

\usepackage[most]{tcolorbox}
\usetikzlibrary{shadows}

\usepackage{amsmath,amsfonts,bm}

\def\eqref#1{equation~\ref{#1}}

\def\1{\bm{1}}

\DeclareMathAlphabet{\mathsfit}{\encodingdefault}{\sfdefault}{m}{sl}
\SetMathAlphabet{\mathsfit}{bold}{\encodingdefault}{\sfdefault}{bx}{n}

\usepackage{hyperref}
\usepackage{url}

\title{%
    \adjustbox{height=2em, valign=m}{\includegraphics[width=2em]{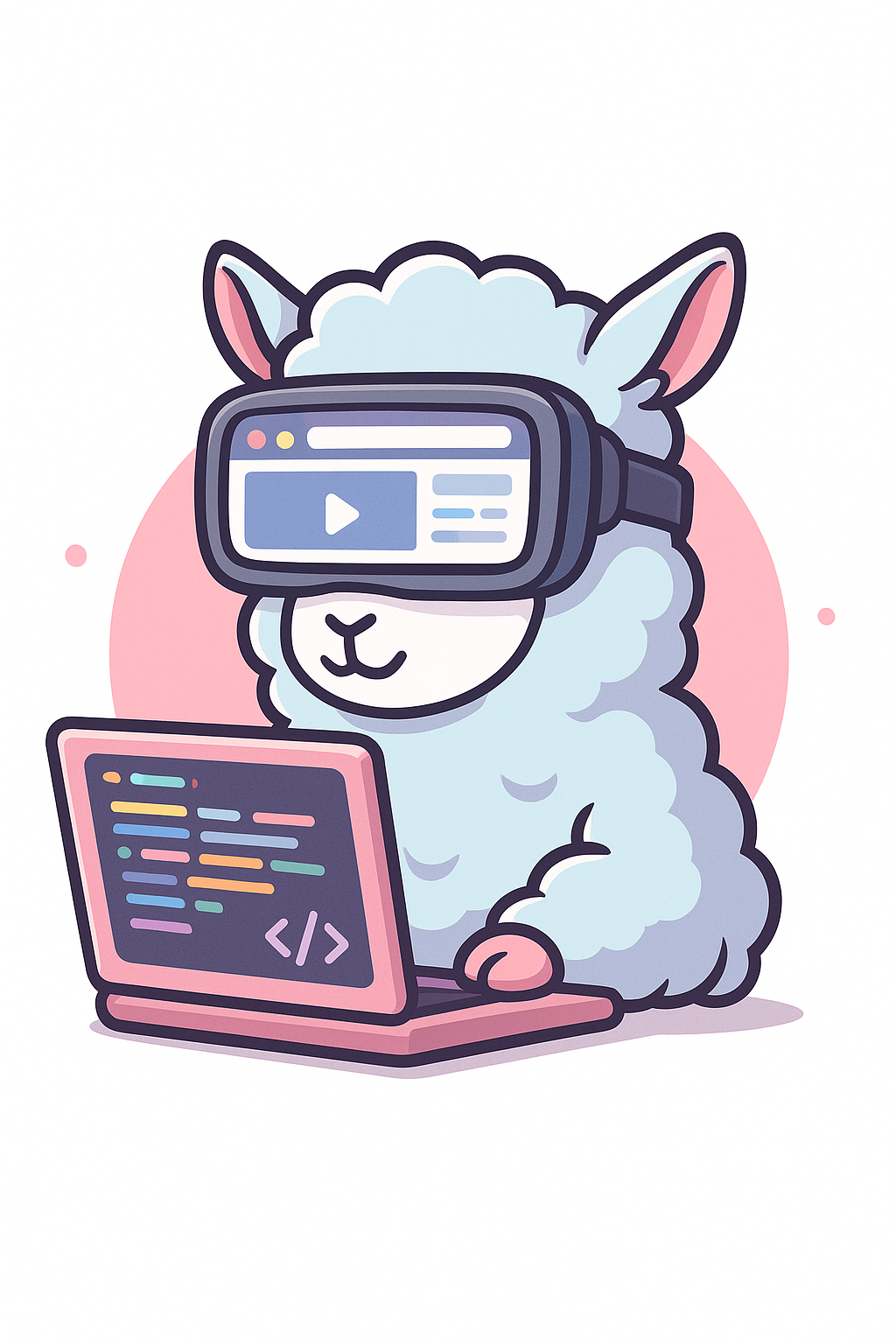}}%
    IWR-Bench: Can LVLMs reconstruct interactive webpage from a user interaction video?
}

\author{
    \textbf{IWR-Bench Team}
}

\usepackage{algorithm}
\usepackage{algorithmicx}
\usepackage{tcolorbox}
\usepackage{xcolor}

\usepackage{listings}

\newcommand{\cmark}{\ding{51}} 
\newcommand{\xmark}{\ding{55}} 
\usepackage{wrapfig}

\lstdefinestyle{prompt}{
    basicstyle=\ttfamily\fontsize{7pt}{8pt}\selectfont,
    frame=none,
    breaklines=true,
    backgroundcolor=\color{lightgray},
    breakatwhitespace=true,
    breakindent=0pt,
    escapeinside={(*@}{@*)},
    numbers=none,
    numbersep=5pt,
    xleftmargin=5pt,
    literate={`}{\textasciigrave}1
}
\tcbset{
  aibox/.style={
    top=10pt,
    colback=white,
    colframe=black,
    colbacktitle=black,
    enhanced,
    center,
    attach boxed title to top left={yshift=-0.1in,xshift=0.15in},
    boxed title style={boxrule=0pt,colframe=white,},
  }
}
\newtcolorbox{AIbox}[2][]{aibox, title=#2,#1}

\iclrfinalcopy
\begin{document}

\maketitle

\begin{abstract}


The webpage-to-code task requires models to understand visual representations of webpages and generate corresponding code.
However, existing benchmarks primarily focus on static screenshot-to-code tasks, thereby overlooking the dynamic interactions fundamental to real-world web applications.
To address this limitation, this paper introduces IWR-Bench, a novel benchmark for evaluating the capabilities of Large Vision-Language Models (LVLMs) in interactive webpage reconstruction from video.
IWR-Bench comprises 113 meticulously curated tasks from 100 real-world websites, with 1,001 actions and featuring diverse interaction complexities (e.g., web games), visual styles, and domains.
Aligning with standard web development practices, each task includes not only user interaction videos but also all crawled static assets (e.g., images, videos).
This benchmark evaluates models on two fundamental challenges: comprehensive multi-modal reasoning to infer interaction logic from video and assets, and advanced code generation to translate this logic into functional code.
An agent-as-a-judge framework with a comprehensive metric system automatically assesses the functional correctness and visual fidelity of generated webpages.
Extensive experiments on 28 LVLMs reveal a significant challenge: the best model achieves an overall score of only 36.35\%, as functional correctness (24.39\% IFS) lags significantly behind visual fidelity (64.25\% VFS).
These results highlight critical limitations in current models' ability to reason about temporal dynamics and synthesize event-driven logic, establishing IWR-Bench as a challenging frontier for vision-language research.
The benchmark and evaluation code will be made publicly available at \textcolor{magenta}{\url{https://github.com/SIGMME/IWR-Bench}}.

\end{abstract}


\begin{figure}[b]
\begin{center}
\includegraphics[width=0.99\linewidth]{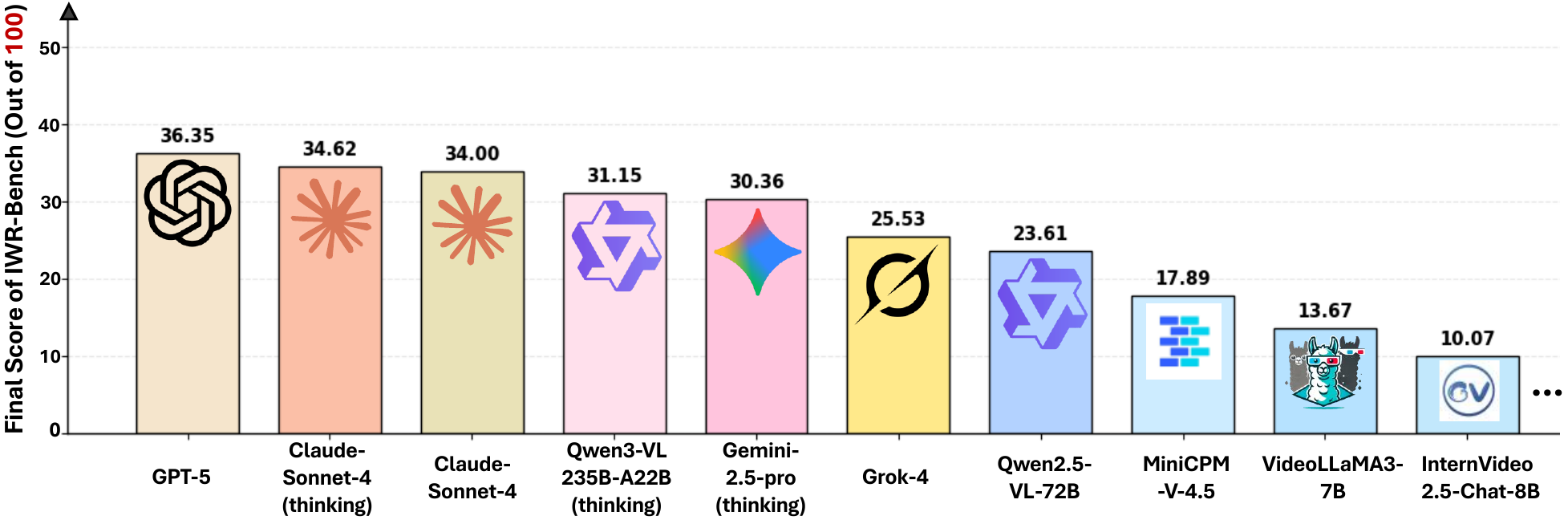}
\end{center}
\caption{Performance of 10 representative models on IWR-Bench. For a comprehensive list of all 28 model results, see Table~\ref{tab:main_results}.}
\label{fig:construction}
\end{figure}

\section{Introduction}

\begin{figure}[h]
\begin{center}
\includegraphics[width=0.99\linewidth]{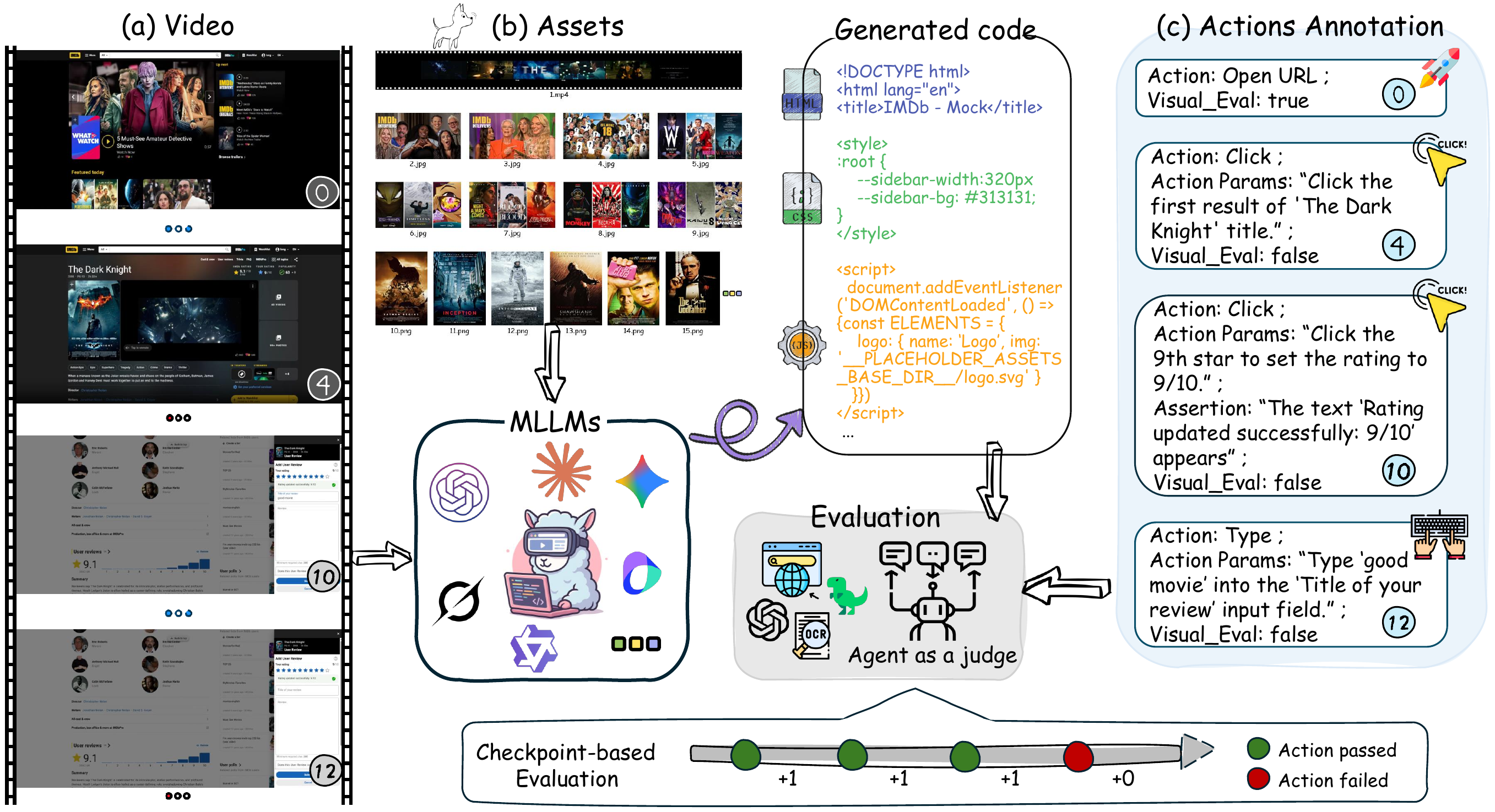}
\end{center}
\caption{Overview of the IWR-Bench task and evaluation. The inputs to the model are (a) a user interaction video and (b) composite images of all static assets sniffed from the webpage. The evaluation employs an agent-as-judge framework~\citep{zhuge2024agent}, where an automated agent assesses the rendered page's interactivity by executing (c) a ground-truth action sequence and its visual fidelity through screenshot comparison.}
\label{fig:overview}
\end{figure}

\begin{figure}[h]
\begin{center}
\includegraphics[width=0.99\linewidth]{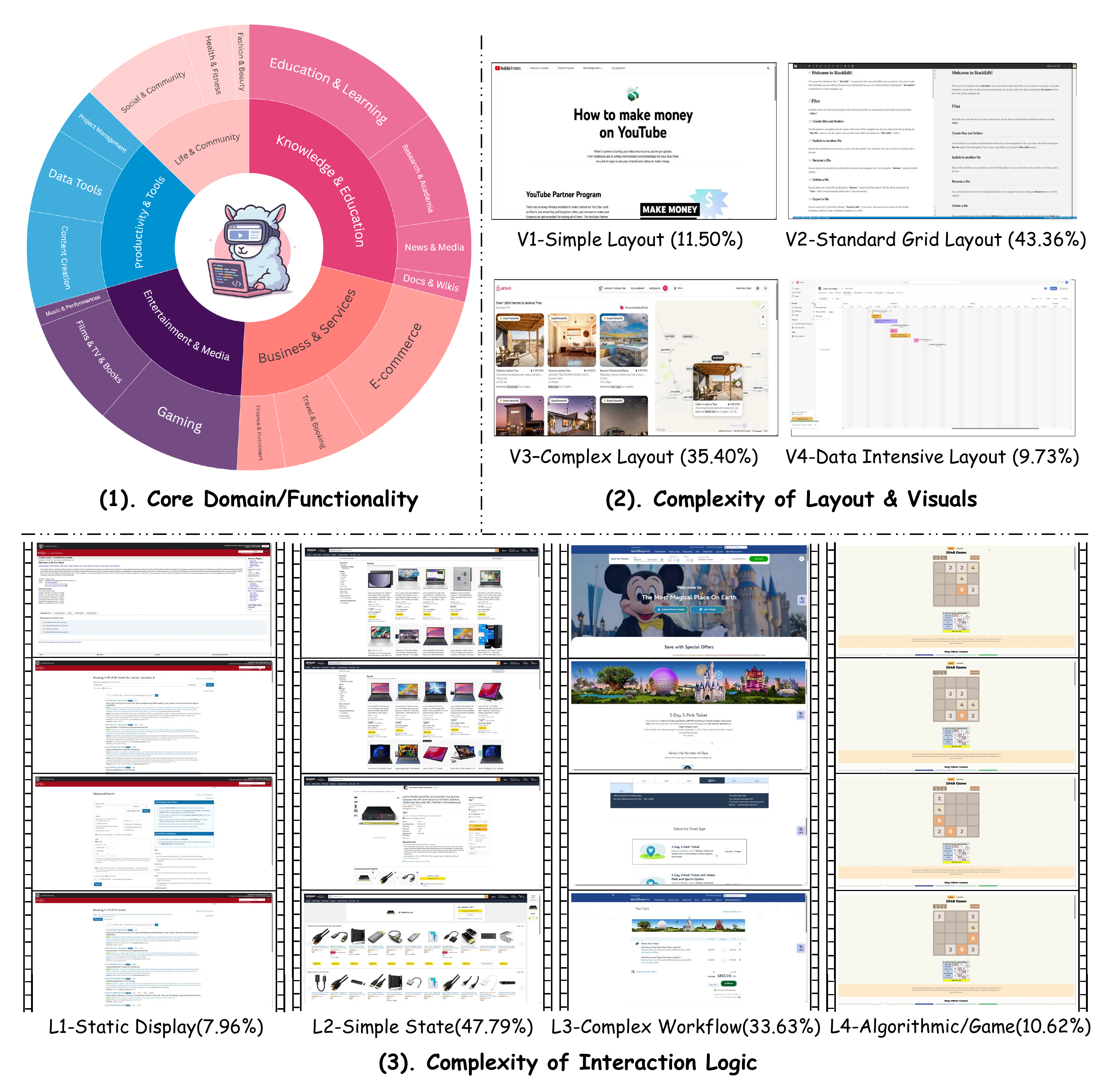}
\end{center}
\caption{An overview of the IWR-Bench taxonomy, which organizes tasks along three orthogonal axes: Domain, Visual Complexity, and Interaction Logic.}
\label{fig:taxonomy}
\end{figure}

Recent advances in Large Vision-Language Models (LVLMs) have unlocked remarkable capabilities in visual understanding and code generation~\citep{openai2025gpt,comanici2025gemini,bai2025qwen2}. State-of-the-art models can now translate a static screenshot of a webpage into corresponding HTML with impressive fidelity~\citep{yun2024web2code,gui2025latcoder}. This nascent success, however, highlights a fundamental limitation of current evaluation methodologies. Existing benchmarks are either confined to static reconstruction (e.g., Design2Code~\citep{si2024design2code}, WebSight~\citep{laurenccon2024unlocking}) or model interactions as single-step, stateless events from image pairs (e.g., Interaction2Code~\citep{xiao2025interaction2code}), while also failing to provide the necessary static assets for reconstruction. This simplified setup falls short of capturing the continuous, stateful workflows and complete resource context characteristic of real-world web applications. The disconnect between demonstrated capabilities and the demands of true interactivity motivates our central research question: \textbf{Can LVLMs reconstruct the dynamic, interactive functionalities of a webpage from observing a user interaction video?}

Reconstructing an interactive webpage from video poses two fundamental challenges. 
The first, \textbf{comprehensive multi-modal perception and reasoning}~\citep{luo2024layoutllm,gupta2023visual,song2025tianyue,deka2017rico,lee2023pix2struct}, is the process of inferring latent interaction logic from dynamic visual evidence. 
This requires a model to ground its temporal understanding of observed interactions in a precise visual comprehension of the resultant UI states. 
A critical facet of this reasoning is robust image matching to associate dynamic elements with their static asset counterparts. 
The second challenge, \textbf{advanced code generation}~\citep{jimenez2024swe,xiao2025interaction2code,li2022competition}, is the translation of this inferred logic into functional code that implements the complex, stateful logic of interactive applications (e.g., web-games like 2048 and Minesweeper).

The construction of a comprehensive benchmark for interactive webpage reconstruction confronts three pivotal challenges. 
The first pertains to ensuring \textbf{Diverse Interaction Coverage}, which necessitates the curation of tasks spanning a broad spectrum of interaction paradigms and visual complexities, while simultaneously adhering to strict standardization for reproducible evaluation. 
The second challenge centers on the establishment of an \textbf{Authentic Task Environment}. Departing from prior benchmarks characterized by incomplete setups or placeholder assets~\citep{jiang2025screencoder,gui2025latcoder}, this requires the meticulous curation of a complete set of authentic resources from live websites. 
Such resources must encompass both static assets, such as images and icons, and dynamic content, such as embedded videos, to faithfully represent real-world development contexts. 
The final challenge lies in the formulation of a \textbf{Robust Automated Evaluation} protocol. Conventional metrics, including pixel-wise similarity, are insufficient for this purpose~\citep{zhang2018unreasonable,caron2021emerging,radford2021learning}, as they cannot appraise functionality. 
An effective protocol must therefore employ programmatic interaction with the generated webpage to ascertain both the functional integrity of its components and the state-wise visual consistency across dynamic transitions.

This paper formalizes the task of Interactive Webpage Reconstruction (IWR) and introduces IWR-Bench, a comprehensive benchmark that addresses these fundamental design challenges. To ensure comprehensive coverage, tasks are taxonomized along orthogonal axes of application domain, visual complexity, and interaction logic, as illustrated in Figure~\ref{fig:taxonomy}. Each task instance, as depicted in Figure~\ref{fig:overview}, then provides the model with (a) an interaction video that captures a complete, stateful workflow, and (b) the full set of crawled static assets. This setup ensures a realistic reconstruction context. Evaluation is conducted via programmatic interaction: an `agent-as-a-judge' executes a ground-truth (c) action sequence to assess the generated webpage's functionality. Performance is quantified by two holistic metrics: the Interactive Functionality Score (IFS), a unified measure of operational and logical correctness, and the Visual Fidelity Score (VFS), a composite metric integrating low-level features with high-level semantic evaluation.

An extensive evaluation on 28 leading LVLMs reveals substantial challenges posed by the IWR task. The top-performing proprietary model, GPT-5, achieves a Final Score of 36.35\%. A clear performance hierarchy is observed, as leading open-source models attain lower scores, and video-specialized models lag even further behind. For the top-performing model, a significant disparity exists between its functional correctness (24.39\% IFS) and visual fidelity (64.25\% VFS). This gap indicates a fundamental limitation across the field: while models can reproduce static layouts with moderate success, their capacity for synthesizing event-driven logic remains severely underdeveloped.

Our key contributions are:

\begin{itemize}[nosep, leftmargin=*]
    \item \textbf{A Benchmark for Interactive Webpage Reconstruction.} We introduce IWR-Bench, the first benchmark to formalize and evaluate Interactive Webpage Reconstruction (IWR) from video. It comprises 113 curated tasks from real-world websites, taxonomized along axes of domain, visual complexity, and interaction logic.
    \item \textbf{A Functionality-Centered Automated Evaluation Protocol.} We develop a robust evaluation protocol that employs a programmatic agent to assess functional correctness by executing ground-truth action sequences. Performance is quantified by two holistic metrics: the Interactive Functionality Score (IFS) and the Visual Fidelity Score (VFS).
    \item \textbf{An Extensive Evaluation and Analysis.} We conduct a comprehensive evaluation of 28 leading LVLMs, establishing strong initial baselines. The results reveal a critical performance gap between visual replication and functional implementation. Further analysis identifies systematic weaknesses in temporal reasoning and logic synthesis, outlining concrete directions for future research.
\end{itemize}

\section{Related Work}

\textbf{Webpage Understanding.}
Webpage understanding evolved from structural analysis based on DOM parsing to a subsequent multimodal perspective that jointly represents a page's visual and textual content~\citep{furuta2023multimodal,burns2023suite,liu2024visualwebbench}. 
Large Vision-Language Models (LVLMs) have advanced webpage understanding by enabling a unified approach where a single model demonstrates strong performance across diverse downstream tasks, indicative of deep comprehension, such as element grounding~\citep{showdown2025} and screen-based question answering~\citep{wang2024webquest,xu2024hierarchical}.
Among these capabilities, generating code from a visual webpage representation is a key task where existing models have demonstrated strong performance~\citep{beltramelli2018pix2code,yun2024web2code,gui2025latcoder}. 
With the enhanced capabilities of LVLMs in handling multiple images or videos~\citep{bai2025qwen2,openai2025gpt,guo2025seed1,comanici2025gemini}, a logical extension of this capability is the generation of interactive webpages, moving beyond static layouts to better mimic real-world applications. 

\textbf{LVLM Benchmarks.}
The development of benchmarks for LVLMs has been driven by the rapid expansion of their capabilities, leading to evaluations of increasing complexity~\citep{jimenez2024swe,yangswe,mialon2023gaia,lu23mathvista}. This progression is evident in the evolution from single-image comprehension to multi-image reasoning and video understanding~\citep{yue2024mmmu,li2023seedbench2,wang2024muirbench,liu2024mmbench,hu2025video,li2024mvbench,fang2024mmbench,fu2025video,ning2023video,chen2024autoeval,yangswe,lu2025webgenbenchevaluatingllmsgenerating}. Concurrently, in the web domain, benchmarks have targeted either webpage understanding or static code generation from a single screenshot~\citep{beltramelli2018pix2code,laurenccon2024unlocking,yun2024web2code,si2024design2code,gui2025latcoder,jiang2025screencoder,awal2025webmmu,xu2025web}, with works like IW-Bench~\citep{guo2024iw} creating more robust evaluation metrics for this task. A recent advancement, Interaction2Code~\citep{xiao2025interaction2code}, extends this by generating code from discrete interaction traces. However, such approaches primarily evaluate single-step, stateless events, rather than the complete, stateful workflows captured in continuous video. Therefore, a critical disconnect exists between model capabilities for dynamic inputs and the benchmarks for interactive web generation.

\section{IWR-BENCH}
\label{sec:benchmark_construction}



\subsection{Task Definition and Structure}

The Interactive Webpage Reconstruction (IWR) task challenges models to generate functional web code from observing user interactions. Formally, given a video $V = \{f_1, ..., f_n\}$ demonstrating user interactions and a set of static assets $A = \{a_1, ..., a_m\}$ from the original webpage, the model must generate code $C$ that reproduces both the visual appearance and interactive behavior observed in $V$. Each task instance in IWR-Bench comprises four key components:


\begin{itemize}[leftmargin=*, topsep=0pt, itemsep=0pt, parsep=0pt]
    \item \textbf{Video Recording:} A screen capture documenting complete user interactions, preserving temporal dynamics and state transitions that define the webpage's behavior.
    \item \textbf{Static Web Assets:} All relevant images, icons, and videos necessary for reconstruction. To prevent models from leveraging prior knowledge based on semantic filenames (e.g., logo.png), all asset filenames are anonymized (e.g., renamed to asset\_001.png)~\citep{agrawal2018don,gurari2018vizwiz}. This forces the model to rely on visual matching and reasoning.
    \item \textbf{Action Trajectory:} A structured sequence $T = \{(a_i, p_i, d_i, v_i, l_i)\}_{i=1}^k$ where each action contains type $a_i$, parameters $p_i$, a natural language description $d_i$, a visual evaluation flag $v_i$, and logical assertions $l_i$ for verification.
    \item \textbf{Checkpoint Screenshots:} Stable-state images $S = \{s_1, ..., s_k\}$ capturing the visual state after each action. This ensures evaluation occurs on fully rendered pages rather than on transitional states.
\end{itemize}

\subsection{Benchmark Construction}
\label{sec:cons}

\begin{figure}[h]
\begin{center}
\includegraphics[width=0.99\linewidth]{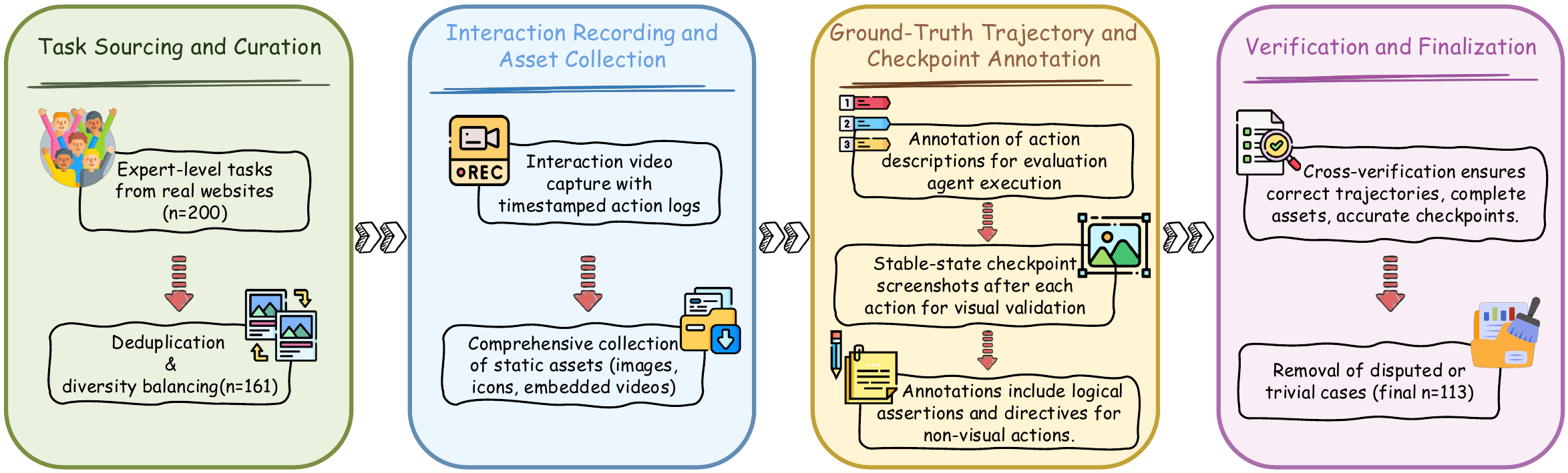}
\end{center}
\caption{The overview of Benchmark construction.}
\label{fig:construction}
\end{figure}

Establishing and maintaining high standards for annotation quality and impartiality is a central design principle in the development of IWR-Bench. 
The process is shown in Figure~\ref{fig:construction}.

\textbf{Task Sourcing and Curation.} The process begins with an initial set of 200 candidate tasks sourced from real-world websites by experts in web development. Each task is defined by a high-level goal and a URL to reflect common usage patterns. Through a rigorous curation process involving deduplication and balancing for diversity across predefined axes, such as domain and complexity (Figure~\ref{fig:taxonomy}), this set is reduced to a high-quality candidate pool of 161 tasks for annotation.

\textbf{Interaction Recording and Asset Collection.}  For each curated task, interactions on the live website are performed by trained annotators and captured as screen recordings, while a browser extension concurrently records the action type $a_i$ and parameters $p_i$ for each action~\citep{yun2024web2code,zhou2023webarena}. In parallel, all relevant static assets, such as images and icons, are collected via automated crawlers and manual inspection.

\textbf{Ground-Truth Trajectory and Checkpoint Annotation.}
The raw recordings and action logs are converted into the final ground-truth representation. For each action, the logged type $a_i$ and parameters $p_i$ are augmented through a three-step annotation process: (1) a natural language description $d_i$ is authored to provide a clear instruction; (2) a visual evaluation flag $v_i$ is assigned, and a corresponding checkpoint screenshot $s_i$ is captured only when this flag is true, signifying a major visual state change; and (3) an optional logical assertion $l_i$ is defined where necessary to programmatically verify functional correctness, such as the appearance of a message or game logic.

\textbf{Verification.} Each annotated task undergoes a two-stage quality assurance process. First, a cross-verification review by a different annotator assesses trajectory correctness, asset completeness, and checkpoint fidelity, with a large model additionally used to verify the accuracy of logical assertions against the ground truth~\citep{zheng2023judging,gou2025mind2web2evaluatingagentic}. All identified discrepancies are rectified. Finally, disputed, ambiguous, or overly trivial tasks are filtered out, leaving a final collection of 113 verified tasks.

\subsection{Taxonomy and Statistics}

\begin{wraptable}{R}{0.48\textwidth}
    \vspace{-1em} 
    \caption{Key Statistics of IWR-Bench.}
    \label{tab:benchmark_stats}
    \centering
    \footnotesize 
    \begin{tabular}{lr}
        \toprule
        \textbf{Statistic} & \textbf{Number} \\
        \midrule
        \multicolumn{2}{l}{\textit{\textbf{Video \& Resolution Statistics}}} \\
        Total Videos & 113 \\
        \quad - Short Videos ($\le$20s) & 25 (22.1\%) \\
        \quad - Medium Videos (20 $\sim$ 60s) & 72 (63.7\%) \\
        \quad - Long Videos ($>$60s) & 16 (14.2\%) \\
        Video Duration (avg/max) & 35.4s / 172.9s \\
        Unique Resolutions & 19 \\
        \quad - Mobile  & 10.62\% \\
        \midrule
        \multicolumn{2}{l}{\textit{\textbf{Evaluation Statistics}}} \\
        Total Actions in Sequences & 1001 \\
        \quad - Visual Evaluation & 620 \\
        \quad - Assertion Checks & 403 \\
        Actions per Video (avg) & 8.9 \\
        \bottomrule
    \end{tabular}
    \vspace{-2em}
\end{wraptable}


IWR-Bench organizes tasks along three orthogonal axes (Figure~\ref{fig:taxonomy}; see Appendix~\ref{sec:app_tasktax} for details): (1) Domain Coverage, which spans 5 major and 16 subcategories such as e-commerce and education to reflect real-world web diversity; (2) Visual Complexity, which scales from minimalist layouts to data-dense dashboards; and (3) Interaction Logic, which progresses from static content display to complex workflows and algorithmic game logic.





Table~\ref{tab:benchmark_stats} presents key statistics. The benchmark includes 113 videos averaging 35.4 seconds, with 1,001 total actions across all sequences. Of these, 620 require visual evaluation and 403 include assertion checks, ensuring comprehensive assessment of both appearance and functionality. Tasks average 8.9 actions, with 10.62\% targeting mobile interfaces, reflecting modern web usage patterns.

\subsection{Comparison to Other Benchmarks}

As detailed in Table~\ref{tab:dataset_comparison}, IWR-Bench addresses a critical gap between webpage reconstruction and video understanding benchmarks. Existing webpage reconstruction benchmarks either focus on static image-to-code tasks (e.g., Pix2Code, WebSight) or model interaction as stateless, single-step events without providing the necessary static assets for reconstruction (e.g., Interaction2Code). Conversely, general video understanding benchmarks (e.g., MVBench) are designed for comprehension tasks like Video QA, not code generation. IWR-Bench overcomes these limitations by using videos of stateful, full-trajectory workflows from live websites. It provides all required assets and employs a robust agent-based protocol to evaluate true interactive correctness.


\begin{table*}[t!]
    \centering 
    \caption{Comparison of IWR-Bench with existing benchmarks. IWR-Bench is unique in its sourcing from live websites, video-based tasks, comprehensive interactive evaluation, and provision of static assets to create a realistic reconstruction task.}
    \label{tab:dataset_comparison} 
    \vspace{-1em}
    \resizebox{\textwidth}{!}{%
    \begin{tabular}{@{}ll l c c c c c@{}} 
        \toprule
        Benchmark & Task Type & Data Source & Videos & \shortstack{Images \\ (Checkpoints)} & \shortstack{Asset \\ Input} & \shortstack{Desktop \\ \& Mobile} & \shortstack{Interactive\\Evaluation} \\
        \midrule
        \multicolumn{8}{l}{\textit{(a) Webpage Reconstruction Benchmarks}} \\
        Pix2Code~\citep{beltramelli2018pix2code}    & Image-to-Code & Synthesized & -- & 1.7K   & \xmark & \cmark & \xmark \\
        DWCG~\citep{yun2024web2code}                & Image-to-Code & Synthesized & -- & 60K    & \xmark & \xmark & \xmark \\
        WebSight~\citep{laurenccon2024unlocking}    & Image-to-Code & Synthesized & -- & 2M  & \xmark & \xmark & \xmark \\
        Design2Code~\citep{si2024design2code}       & Image-to-Code & C4 & -- & 484    & \xmark & \xmark & \xmark \\
        CC-HARD~\citep{gui2025latcoder}             & Image-to-Code & C4 & -- & 128    & \xmark & \xmark & \xmark \\
        ScreenCoder~\citep{jiang2025screencoder}     & Image-to-Code & Live Websites & -- & 3K     & \xmark & \xmark & \xmark \\
        Interaction2Code~\citep{xiao2025interaction2code} & Images-to-Code & C4 \& GitHub & -- & 374 & \xmark & \xmark & \cmark \small{(Single-step)} \\
        \midrule
        \multicolumn{8}{l}{\textit{(b) Video Understanding Benchmarks}} \\
        MMBench-Video~\citep{fang2024mmbench} & Video QA & - & 609 & -- & -- & -- & -- \\
        MVBench~\citep{li2024mvbench}           & Video QA & - & 20K & -- & -- & -- & -- \\
        Video-MME~\citep{fu2025video}         & Video QA & - & 900 & -- & -- & -- & -- \\
        Video-MMMU~\citep{hu2025video}        & Video QA & - & 300 & -- & -- & -- & -- \\
        \midrule 
        \textbf{IWR-Bench (Ours)} & \textbf{Video-to-Code} & \textbf{Live Websites} & \textbf{113} & \textbf{620*} & \cmark & \cmark & \cmark \small{(Full Trajectory)} \\
        \bottomrule
    \end{tabular}%
    }
    \begin{minipage}{\textwidth}
    \vspace{1mm}
    \small * These are images used for evaluating visual fidelity across interaction states.
    \end{minipage}
    \vspace{-2.2em}
\end{table*}

\section{Evaluation and Metrics}
\label{sec:evaluation}

\subsection{Evaluation Protocol}
\label{sec:protocol}

The evaluation of generated code $C$ is conducted using a deterministic executor built upon the \textit{browser-use} library~\citep{browser-use}. This executor programmatically interacts with the rendered webpage by sequentially executing each pre-defined action $a_i$ from the ground-truth action trajectory $T$. This design isolates the evaluation to code execution and removes any dependency on high-level task planning, thereby ensuring a stable and reproducible protocol.

The evaluation of the trajectory proceeds step-by-step. At each step $i$, the action $a_i$ is attempted. The action is considered a failure under two conditions: (1) it is operationally infeasible (e.g., a target element is not found), or (2) its corresponding logical assertions $l_i$ are not satisfied. 

For logical assertion verification, an MLLM judge, specifically \textbf{Gemini-2.5-Pro}~\citep{comanici2025gemini}, is employed to analyze screenshots of the page state before and after an action to determine its correctness. The prompt for this judge is detailed in Appendix~\ref{sec:app_prompts}. Upon the successful completion of an action, a new screenshot is captured. If the visual evaluation flag $v_i$ for this step is true, the new screenshot undergoes a visual fidelity assessment. This assessment is based on a composite score that integrates OCR-based text similarity~\citep{cui2025paddleocr}, DINO-based structural similarity~\citep{oquab2023dinov2}, and a high-level evaluation also conducted by \textbf{Gemini-2.5-Pro} (see Appendix~\ref{sec:app_prompts} for the prompt). Actions where $v_i$ is false, which typically involve insignificant or stochastic visual changes, are omitted from this visual assessment phase.

\subsection{Metrics}
\label{sec:metrics}
Model performance is quantified through a hierarchy of metrics designed to measure functional correctness, visual fidelity, and overall task completion.

\paragraph{Interactive Functionality Score (IFS).}
This metric measures a model's ability to generate functionally correct code. An action $a_i$ from the trajectory $T$ is considered successful if and only if it executes without operational errors and all associated logical assertions $l_i$ are satisfied, as determined by the protocol in Section~\ref{sec:protocol}. The IFS is defined as the ratio of successfully completed actions ($N_{\text{succ}}$) to the total number of actions ($N_{\text{total}}$).
\begin{equation} \label{eq:ifs}
    \text{IFS} = \frac{N_{\text{succ}}}{N_{\text{total}}}
\end{equation}

\paragraph{Visual Fidelity Score (VFS).}
The VFS assesses the visual quality of the rendered user interface. This score is computed exclusively over checkpoints that were successfully reached and have the visual evaluation flag enabled ($v_i = \text{true}$). Let $I_{v, \text{succ}}$ be the set of indices for these qualifying checkpoints. The score for each checkpoint $i \in I_{v, \text{succ}}$ is a weighted combination of two components: a \textit{Low-level Visual Score} ($S_{\text{LVS}, i}$), which averages an OCR-based Levenshtein similarity and a DINO-based cosine similarity, and a \textit{High-level Visual Score} ($S_{\text{HVS}, i}$), which is a holistic assessment from the MLLM judge. The final VFS is the macro-average of these checkpoint scores. The weight $w$ is set to 0.5 based on validation studies (Section~\ref{sec:validation}).
\begin{equation} \label{eq:vfs_final}
\text{VFS} = \frac{1}{|I_{v, \text{succ}}|} \sum_{i \in I_{v, \text{succ}}} \left( w \cdot S_{\text{LVS}, i} + (1-w) \cdot S_{\text{HVS}, i} \right)
\end{equation}

\paragraph{Final Score.}

The Final Score is defined by combining the IFS and VFS with fixed weights. For steps where actions cannot be executed, no images are available to compute visual similarity scores. An alternative weighting scheme based on the ratio of successful ($N_{\text{succ}}$) to total ($N_{\text{total}}$) steps was explored, but it proved ineffective for differentiating model performance. Therefore, a simple weighted combination is adopted with the weighting factor $\alpha$ set to 0.7 (see Section~\ref{sec:validation}). By assigning substantial weight to the IFS component, the impact of unreachable states on overall evaluation is appropriately reflected. All reported scores are macro-averaged across the entire benchmark.
\begin{equation} \label{eq:final_score}
\text{Final Score} = \alpha \cdot \text{IFS} + (1-\alpha) \cdot \text{VFS}
\end{equation}

\section{Experiments}

\subsection{Evaluation Setup}
\label{sec:eval_setup}
\textbf{Evaluation Models.} The evaluation is conducted on a diverse set of 28 leading Large Vision-Language Models (LVLMs) to establish a comprehensive performance baseline on IWR-Bench. This selection encompasses both proprietary and open-source models, as well as specialized video understanding models. The full list of evaluated models and their performance is detailed in Table~\ref{tab:main_results}.

\textbf{Implementation Details.} For each task in IWR-Bench, models are provided with the user interaction video and a composite image of all crawled static assets. To accommodate models without native video support, each video is sampled at 1 fps, with the number of frames capped at 64. Videos exceeding 64 seconds are uniformly downsampled to meet this limit. The video (or its sampled frames) and the composite image are arranged as a sequential, multi-image input. The task is to generate a single, self-contained HTML file that integrates all necessary CSS and JavaScript to replicate the observed webpage. All other inference parameters utilize the default settings recommended by the model providers. The complete prompt templates are detailed in Appendix~\ref{sec:app_prompts}.

\subsection{Main Results}


\definecolor{Gray}{gray}{0.9}

\begin{table*}[t]
\centering
\caption{Main evaluation results on IWR-Bench. Models are grouped by category and sorted by Final Score. Reasoning-enhanced ('thinking') model variants are highlighted in gray. The best result in each column is \textbf{bolded}, and the second-best is \underline{underlined}.}
\label{tab:main_results}
\resizebox{\textwidth}{!}{%
\begin{tabular}{lccccc}
\toprule
\textbf{Model} & 
\makecell{\textbf{L}ow-level \\ \textbf{V}isual \\ \textbf{S}core} & 
\makecell{\textbf{H}igh-level \\ \textbf{V}isual \\ \textbf{S}core} & 
\makecell{\textbf{V}isual \\ \textbf{F}idelity \\ \textbf{S}core} & 
\makecell{\textbf{I}nteractive \\ \textbf{F}unctionality \\ \textbf{S}core} & 
\makecell{\textbf{F}inal \\ \textbf{S}core} \\
\midrule
\multicolumn{6}{l}{\textit{Proprietary MLLMs}} \\
\quad GPT-5~\citep{openai2025gpt} & \textbf{68.29} & \textbf{60.21} & \textbf{64.25} & \textbf{24.39} & \textbf{36.35} \\
\rowcolor{Gray}
\quad Claude-Sonnet-4 (thinking)~\citep{claudesonnet4} & 64.90 & 55.51 & 60.20 & \underline{23.65} & \underline{34.62} \\
\rowcolor{Gray}
\quad Claude-Opus-4 (thinking)~\citep{claudeopus} & 63.53 & 53.80 & 58.67 & 23.61 & 34.13 \\
\quad Doubao-seed-1.6~\citep{bytedancese} & \underline{65.95} & 55.62 & 60.79 & 22.55 & 34.02 \\
\quad Claude-Sonnet-4~\citep{claudesonnet4} & 65.75 & \underline{56.92} & \underline{61.34} & 22.29 & 34.00 \\
\quad Claude-Opus-4~\citep{claudeopus} & 65.23 & 55.13 & 60.18 & 21.83 & 33.33 \\
\quad GPT-5-mini~\citep{openai2025gpt} & 63.36 & 50.25 & 56.81 & 23.18 & 33.27 \\
\quad GPT-4.1~\citep{openaigpt41} & 63.07 & 54.63 & 58.85 & 20.48 & 31.99 \\
\rowcolor{Gray}
\quad Gemini-2.5-Pro (thinking)~\citep{comanici2025gemini} & 54.52 & 46.83 & 50.67 & 21.65 & 30.36 \\
\quad Gemini-2.5-Pro~\citep{comanici2025gemini} & 57.46 & 48.91 & 53.18 & 20.51 & 30.31 \\
\quad GPT-4o (latest)~\citep{hurst2024gpt} & 63.39 & 51.71 & 57.55 & 17.55 & 29.55 \\
\quad Gemini-2.5-Flash~\citep{comanici2025gemini} & 47.53 & 37.75 & 42.64 & 19.88 & 26.71 \\
\quad GPT-5-nano~\citep{openai2025gpt} & 53.49 & 35.70 & 44.59 & 18.17 & 26.10 \\
\quad Grok-4~\citep{grok_4} & 48.95 & 30.54 & 39.74 & 19.44 & 25.53 \\
\quad GPT-4o (0806)~\citep{hurst2024gpt} & 54.03 & 39.83 & 46.93 & 15.87 & 25.19 \\
\quad Doubao-seed-1.6-flash~\citep{bytedancese} & 45.49 & 32.06 & 38.78 & 16.34 & 23.07 \\
\quad Gemini-2.5-Flash-Lite~\citep{comanici2025gemini} & 28.95 & 19.05 & 24.00 & 13.29 & 16.50 \\
\midrule
\multicolumn{6}{l}{\textit{Open-Source MLLMs}} \\
\rowcolor{Gray}
\quad Qwen3-VL (thinking)~\citep{qwen3-vl} & 58.55 & 46.13 & 52.34 & 22.07 & 31.15 \\
\quad Qwen2.5-VL-72B~\citep{bai2025qwen2} & 47.83 & 28.25 & 38.04 & 17.42 & 23.61 \\
\quad Qwen2.5-VL-32B~\citep{bai2025qwen2} & 39.36 & 23.30 & 31.33 & 16.50 & 20.95 \\
\quad Keye-VL-1.5-8B~\citep{yang2025kwaikeyevl15technical} & 30.81 & 15.49 & 23.15 & 16.06 & 18.18 \\
\quad MiniCPM-V-4.5~\citep{yu2025minicpmv45cookingefficient} & 31.18 & 15.41 & 23.29 & 15.58 & 17.89 \\
\quad Qwen2.5-VL-7B~\citep{bai2025qwen2} & 28.92 & 12.20 & 20.56 & 13.28 & 15.47 \\
\rowcolor{Gray}
\quad Kimi-VL (thinking)~\citep{kimiteam2025kimivltechnicalreport} & 26.18 & 12.23 & 19.20 & 12.04 & 14.19 \\
\quad Mimo-VL-7B~\citep{coreteam2025mimovltechnicalreport} & 23.28 & 4.99 & 14.14 & 10.57 & 11.64 \\
\quad GLM-4.5V~\citep{vteam2025glm45vglm41vthinkingversatilemultimodal} & 16.31 & 10.52 & 13.41 & 10.11 & 11.10 \\
\midrule
\multicolumn{6}{l}{\textit{Open-Source Video-Specialized LMs}} \\
\quad VideoLLaMA3-7B~\citep{zhang2025videollama} & 31.29 & 11.86 & 21.58 & 10.29 & 13.67 \\
\quad InternVideo-2.5-Chat-8B~\citep{wang2025internvideo2} & 17.27 & 3.33 & 10.30 & 9.97 & 10.07 \\
\bottomrule
\end{tabular}%
}
\end{table*}

The comprehensive evaluation results on IWR-Bench are presented in Table~\ref{tab:main_results}. The findings reveal a clear performance landscape, highlighting the substantial difficulty of the task and surfacing several key observations regarding current model capabilities, with case studies provided in Appendix~\ref{sec:app_case_study}.

\textbf{A Clear Performance Hierarchy Is Observed Across Model Categories.}
The results on IWR-Bench show a pronounced performance stratification across model categories. Proprietary multimodal large language models are positioned in the upper echelon, with GPT-5 obtaining the highest Final Score (36.35). This is followed by a competitive cluster that includes Claude-Sonnet-4 (thinking) (34.62), Claude-Opus-4 (thinking) (34.13), Doubao-seed-1.6 (34.02), and Claude-Sonnet-4 (34.00). The top-performing open-source model, Qwen3-VL (thinking), has a score of 31.15. This score is lower than that of the leading proprietary group but surpasses several mid-tier proprietary entries, such as GPT-4o (latest) (29.55). At the lower end of the performance spectrum, video-specialized models like VideoLLaMA3-7B (13.67) and InternVideo-2.5-Chat-8B (10.07) are found. This hierarchy indicates that general multimodal reasoning and code generation capabilities are more critical for success on IWR-Bench than specialized video-processing architectures.

\textbf{Interactive Functionality Remains the Primary Performance Bottleneck.}
A substantial performance gap exists between static visual replication and dynamic functionality implementation. This gap is reflected in the consistently higher Visual Fidelity Scores (VFS) compared to the Interactive Functionality Scores (IFS). For instance, GPT-5 obtains the highest visual metrics (LVS 68.29, HVS 60.21, VFS 64.25), yet its corresponding IFS is only 24.39. A similar pattern is observed for Claude-Sonnet-4, which has the second-highest VFS (61.34) but an IFS of only 22.29. The difficulty of this task is further underscored by the low absolute IFS values, with the highest score remaining below 25, highlighting that interactive webpage reconstruction is a largely unsolved problem.


\textbf{Reasoning Enhancement Provides Consistent but Moderate Gains.}
Consistent but moderate performance improvements are observed when using reasoning-enhanced inference. For instance, the "thinking" variant of Claude-Sonnet-4 shows higher performance in both Final Score (34.62 vs. 34.00) and IFS (23.65 vs. 22.29). A similar trend is noted for Claude-Opus-4 (Final 34.13 vs. 33.33; IFS 23.61 vs. 21.83) and Gemini-2.5-Pro (Final 30.36 vs. 30.31; IFS 21.65 vs. 20.51). This evidence indicates that while enhanced reasoning acts as a useful refinement, the base model's capability remains the primary factor determining the performance ceiling on IWR-Bench.

\subsection{Performance Analysis Across Task Dimensions}

A fine-grained analysis (detailed in Appendix~\ref{appendix:detailed_results}) reveals distinct performance patterns. The synthesis of event-driven functionality is the primary bottleneck, evidenced by a sharp performance drop from static (L1) to interactive (L2-L4) tasks (Table~\ref{tab:interaction_breakdown_final}). Models also struggle with highly structured layouts (Table~\ref{tab:visual_breakdown_final}). Performance varies by domain, with relative strength in "Entertainment \& Media" (Table~\ref{tab:domain_breakdown_final}), pointing to structured code generation and state management as key research directions.

\subsection{Validation of the Evaluation Protocol}
\label{sec:validation}
The robustness and reliability of the evaluation protocol are validated through a rigorous, two-part analysis that addresses both the metric parameters and the agent-as-a-judge methodology~\citep{zheng2023judging,gou2025mind2web2evaluatingagentic,maaz2023video}. 
First, the weighting coefficients ($w$ and $\alpha$) for the scoring metrics are determined through a human alignment study, with the detailed procedure and results presented in Appendix~\ref{sec:app_metric_par}.
Second, the agent-as-a-judge framework is validated through a multi-stage process. 
This process includes a meticulous cross-verification of annotated action trajectories (Section~\ref{sec:cons}), automated verification of logical assertions using an MLLM-based judge\citep{comanici2025gemini}, and manual inspection of the agent's operational fidelity.
For the manual inspection, three PhD students observed the agent's execution on 100 randomly sampled, model-generated webpages, with the browser's headless mode disabled to compare on-screen behavior against evaluation logs.
Failures in the agent's evaluation were observed in only three instances. 
These issues typically stemmed from ambiguous element descriptors (e.g., buttons with identical names), which required a more precise locator (\texttt{d\_i}). 
All identified discrepancies were subsequently rectified.

\section{Conclusion}

This paper introduces IWR-Bench, the first benchmark designed to evaluate Interactive Webpage Reconstruction from video. Through an automated agent-as-a-judge evaluation protocol, performance is quantified using two metrics: the IFS and the VFS. Comprehensive evaluations on 28 LVLMs reveal a stark disparity between visual replication and functional implementation. While models achieve moderate success in reconstructing static appearance (VFS), their ability to generate correct, event-driven logic remains critically limited, as shown by low IFS scores across the board. This finding indicates that the primary bottleneck for current models is not visual understanding but the synthesis of complex interaction logic. IWR-Bench thus establishes a challenging new frontier for vision-language research, highlighting the need for future work to focus on temporal reasoning, dynamic asset binding, and robust code synthesis to create truly functional web applications.

\clearpage

\section{Contributors}


{
\renewcommand{\thefootnote}{*}
\textbf{Core Contributors\footnote{Equal contribution.}}
}
\begin{itemize}[noitemsep, topsep=0.3em, partopsep=0pt]
    \item Yang Chen, \textit{Shanghai AI Lab, Zhejiang University} \hspace{1em} \textbf{zjucheny@gmail.com}
    \item Minghao Liu, \textit{2077AI, M-A-P}
    \item Yufan Shen, \textit{Shanghai AI Lab}
    \item Yunwen Li, \textit{Chinese University of Hong Kong(shenzhen), M-A-P}
    \item Tianyuan Huang, \textit{Zhejiang University}
    \item Xinyu Fang, \textit{Shanghai AI Lab, Zhejiang University}
\end{itemize}

\textbf{Contributors}
\begin{itemize}[noitemsep, topsep=0.3em, partopsep=0pt]
    \item Tianyu Zheng, \textit{M-A-P}
    \item Wenxuan Huang, \textit{Chinese University of Hong Kong}
    \item Cheng Yang, \textit{Shanghai AI Lab, Central South University}
    \item Daocheng Fu, \textit{Shanghai AI Lab, Fudan University}
    \item Jianbiao Mei, \textit{Shanghai AI Lab, Zhejiang University}
    \item Rong Wu, \textit{Shanghai AI Lab, Zhejiang University}
    \item Yunfei Zhao, \textit{Stanford University, 2077AI}
\end{itemize}

\textbf{Advisors}
\begin{itemize}[noitemsep, topsep=0.3em, partopsep=0pt]
    \item Licheng Wen, \textit{Shanghai AI Lab}
    \item Xuemeng Yang, \textit{Shanghai AI Lab}
    \item Song Mao, \textit{Shanghai AI Lab}
    \item Qunshu Lin, \textit{2077AI}
    \item Zhi Yu, \textit{Zhejiang University}
    \item Yongliang Shen, \textit{Zhejiang University}
    \item Yu Qiao, \textit{Shanghai AI Lab}
\end{itemize}

\textbf{Corresponding Authors}
\begin{itemize}[noitemsep, topsep=0.3em, partopsep=0pt]
    \item Yufan Shen, \textit{Shanghai AI Lab} \hspace{1em}  \textbf{shenyfzju@gmail.com}
    \item Botian Shi, \textit{Shanghai AI Lab, Shanghai Innovation Institute} \textbf{shibotian@pjlab.org.cn}
\end{itemize}

\textbf{Project Leader}
\begin{itemize}[noitemsep, topsep=0.3em, partopsep=0pt]
    \item Yufan Shen, \textit{Shanghai AI Lab} 
\end{itemize}

\clearpage

\bibliography{iwr}

@misc{showdown2025,
  title={The Showdown Computer Control Evaluation Suite},
  author={General Agents Team},
  year={2025},
  url={https://github.com/generalagents/showdown},
}

@article{furuta2023multimodal,
  title={Multimodal web navigation with instruction-finetuned foundation models},
  author={Furuta, Hiroki and Lee, Kuang-Huei and Nachum, Ofir and Matsuo, Yutaka and Faust, Aleksandra and Gu, Shixiang Shane and Gur, Izzeddin},
  journal={arXiv preprint arXiv:2305.11854},
  year={2023}
}

@article{burns2023suite,
  title={A suite of generative tasks for multi-level multimodal webpage understanding},
  author={Burns, Andrea and Srinivasan, Krishna and Ainslie, Joshua and Brown, Geoff and Plummer, Bryan A and Saenko, Kate and Ni, Jianmo and Guo, Mandy},
  journal={arXiv preprint arXiv:2305.03668},
  year={2023}
}

@article{liu2024visualwebbench,
  title={Visualwebbench: How far have multimodal llms evolved in web page understanding and grounding?},
  author={Liu, Junpeng and Song, Yifan and Lin, Bill Yuchen and Lam, Wai and Neubig, Graham and Li, Yuanzhi and Yue, Xiang},
  journal={arXiv preprint arXiv:2404.05955},
  year={2024}
}

@inproceedings{xu2024hierarchical,
  title={Hierarchical multimodal pre-training for visually rich webpage understanding},
  author={Xu, Hongshen and Chen, Lu and Zhao, Zihan and Ma, Da and Cao, Ruisheng and Zhu, Zichen and Yu, Kai},
  booktitle={Proceedings of the 17th ACM International Conference on Web Search and Data Mining},
  pages={864--872},
  year={2024}
}

@article{wang2024webquest,
  title={Webquest: A benchmark for multimodal qa on web page sequences},
  author={Wang, Maria and Sunkara, Srinivas and Baechler, Gilles and Lin, Jason and Zhu, Yun and Zubach, Fedir and Shu, Lei and Chen, Jindong},
  journal={arXiv preprint arXiv:2409.13711},
  year={2024}
}

@inproceedings{beltramelli2018pix2code,
  title={pix2code: Generating code from a graphical user interface screenshot},
  author={Beltramelli, Tony},
  booktitle={Proceedings of the ACM SIGCHI symposium on engineering interactive computing systems},
  pages={1--6},
  year={2018}
}

@article{jiang2025screencoder,
  title={ScreenCoder: Advancing Visual-to-Code Generation for Front-End Automation via Modular Multimodal Agents},
  author={Jiang, Yilei and Zheng, Yaozhi and Wan, Yuxuan and Han, Jiaming and Wang, Qunzhong and Lyu, Michael R and Yue, Xiangyu},
  journal={arXiv preprint arXiv:2507.22827},
  year={2025}
}

@article{yun2024web2code,
  title={Web2code: A large-scale webpage-to-code dataset and evaluation framework for multimodal llms},
  author={Yun, Sukmin and Thushara, Rusiru and Bhat, Mohammad and Wang, Yongxin and Deng, Mingkai and Wang, Jinhong and Tao, Tianhua and Li, Junbo and Li, Haonan and Nakov, Preslav and others},
  journal={Advances in neural information processing systems},
  volume={37},
  pages={112134--112157},
  year={2024}
}

@inproceedings{gurari2018vizwiz,
  title={Vizwiz grand challenge: Answering visual questions from blind people},
  author={Gurari, Danna and Li, Qing and Stangl, Abigale J and Guo, Anhong and Lin, Chi and Grauman, Kristen and Luo, Jiebo and Bigham, Jeffrey P},
  booktitle={Proceedings of the IEEE conference on computer vision and pattern recognition},
  pages={3608--3617},
  year={2018}
}

@article{zhou2023webarena,
  title={Webarena: A realistic web environment for building autonomous agents},
  author={Zhou, Shuyan and Xu, Frank F and Zhu, Hao and Zhou, Xuhui and Lo, Robert and Sridhar, Abishek and Cheng, Xianyi and Ou, Tianyue and Bisk, Yonatan and Fried, Daniel and others},
  journal={arXiv preprint arXiv:2307.13854},
  year={2023}
}

@inproceedings{agrawal2018don,
  title={Don't just assume; look and answer: Overcoming priors for visual question answering},
  author={Agrawal, Aishwarya and Batra, Dhruv and Parikh, Devi and Kembhavi, Aniruddha},
  booktitle={Proceedings of the IEEE conference on computer vision and pattern recognition},
  pages={4971--4980},
  year={2018}
}

@article{li2022competition,
  title={Competition-level code generation with alphacode},
  author={Li, Yujia and Choi, David and Chung, Junyoung and Kushman, Nate and Schrittwieser, Julian and Leblond, R{\'e}mi and Eccles, Tom and Keeling, James and Gimeno, Felix and Dal Lago, Agustin and others},
  journal={Science},
  volume={378},
  number={6624},
  pages={1092--1097},
  year={2022},
  publisher={American Association for the Advancement of Science}
}

@inproceedings{lee2023pix2struct,
  title={Pix2struct: Screenshot parsing as pretraining for visual language understanding},
  author={Lee, Kenton and Joshi, Mandar and Turc, Iulia Raluca and Hu, Hexiang and Liu, Fangyu and Eisenschlos, Julian Martin and Khandelwal, Urvashi and Shaw, Peter and Chang, Ming-Wei and Toutanova, Kristina},
  booktitle={International Conference on Machine Learning},
  pages={18893--18912},
  year={2023},
  organization={PMLR}
}

@inproceedings{deka2017rico,
  title={Rico: A mobile app dataset for building data-driven design applications},
  author={Deka, Biplab and Huang, Zifeng and Franzen, Chad and Hibschman, Joshua and Afergan, Daniel and Li, Yang and Nichols, Jeffrey and Kumar, Ranjitha},
  booktitle={Proceedings of the 30th annual ACM symposium on user interface software and technology},
  pages={845--854},
  year={2017}
}

@article{laurenccon2024unlocking,
  title={Unlocking the conversion of web screenshots into html code with the websight dataset},
  author={Lauren{\c{c}}on, Hugo and Tronchon, L{\'e}o and Sanh, Victor},
  journal={arXiv preprint arXiv:2403.09029},
  year={2024}
}

@inproceedings{gui2025latcoder,
  title={LaTCoder: Converting Webpage Design to Code with Layout-as-Thought},
  author={Gui, Yi and Li, Zhen and Zhang, Zhongyi and Wang, Guohao and Lv, Tianpeng and Jiang, Gaoyang and Liu, Yi and Chen, Dongping and Wan, Yao and Zhang, Hongyu and others},
  booktitle={Proceedings of the 31st ACM SIGKDD Conference on Knowledge Discovery and Data Mining V. 2},
  pages={721--732},
  year={2025}
}

@article{bai2025qwen2,
  title={Qwen2. 5-vl technical report},
  author={Bai, Shuai and Chen, Keqin and Liu, Xuejing and Wang, Jialin and Ge, Wenbin and Song, Sibo and Dang, Kai and Wang, Peng and Wang, Shijie and Tang, Jun and others},
  journal={arXiv preprint arXiv:2502.13923},
  year={2025}
}

@misc{yang2025kwaikeyevl15technical,
      title={Kwai Keye-VL 1.5 Technical Report}, 
      author={Biao Yang and Bin Wen and Boyang Ding and Changyi Liu and Chenglong Chu and Chengru Song and Chongling Rao and others},
      year={2025},
      eprint={2509.01563},
      archivePrefix={arXiv},
      primaryClass={cs.CV},
      url={https://arxiv.org/abs/2509.01563}, 
}

@article{zhang2025videollama,
  title={Videollama 3: Frontier multimodal foundation models for image and video understanding},
  author={Zhang, Boqiang and Li, Kehan and Cheng, Zesen and Hu, Zhiqiang and Yuan, Yuqian and Chen, Guanzheng and Leng, Sicong and Jiang, Yuming and Zhang, Hang and Li, Xin and others},
  journal={arXiv preprint arXiv:2501.13106},
  year={2025}
}

@misc{openaigpt41,
  author       = {{OpenAI}},
  title        = {{gpt-4-1}},
  howpublished = {\url{https://openai.com/index/gpt-4-1/}},
  year         = {2025},
  month        = {April},
  day          = {14}
}

@misc{qwen3-vl,
  author       = {{QwenTeam}},
  title        = {{qwen3-vl}},
  howpublished = {\url{https://qwen.ai/blog?id=99f0335c4ad9ff6153e517418d48535ab6d8afef&from=research.latest-advancements-list}},
  year         = {2025},
  month        = {September},
  day          = {23}
}

@misc{yu2025minicpmv45cookingefficient,
      title={MiniCPM-V 4.5: Cooking Efficient MLLMs via Architecture, Data, and Training Recipe}, 
      author={Tianyu Yu and Zefan Wang and Chongyi Wang and Fuwei Huang and Wenshuo Ma and Zhihui He and Tianchi Cai and Weize Chen and Yuxiang Huang and Yuanqian Zhao and Bokai Xu and Junbo Cui and Yingjing Xu and Liqing Ruan and Luoyuan Zhang and Hanyu Liu and Jingkun Tang and Hongyuan Liu and Qining Guo and Wenhao Hu and Bingxiang He and Jie Zhou and Jie Cai and Ji Qi and Zonghao Guo and Chi Chen and Guoyang Zeng and Yuxuan Li and Ganqu Cui and Ning Ding and Xu Han and Yuan Yao and Zhiyuan Liu and Maosong Sun},
      year={2025},
      eprint={2509.18154},
      archivePrefix={arXiv},
      primaryClass={cs.LG},
      url={https://arxiv.org/abs/2509.18154}, 
}

@misc{coreteam2025mimovltechnicalreport,
      title={MiMo-VL Technical Report}, 
      author={Core Team and Zihao Yue and Zhenru Lin and Yifan Song and Weikun Wang and Shuhuai Ren and Shuhao Gu and Shicheng Li and Peidian Li and Liang Zhao and Lei Li and Kainan Bao and Hao Tian and Hailin Zhang and Gang Wang and others},
      year={2025},
      eprint={2506.03569},
      archivePrefix={arXiv},
      primaryClass={cs.CL},
      url={https://arxiv.org/abs/2506.03569}, 
}

@misc{claudesonnet4,
  author       = {{anthropic}},
  title        = {{claude-sonnet-4}},
  howpublished = {\url{https://www.anthropic.com/claude/sonnet}},
  year         = {2025}
}

@misc{claudeopus,
  author       = {{anthropic}},
  title        = {{claude-opus}},
  howpublished = {\url{https://www.anthropic.com/claude/opus}},
  year         = {2025}
}

@misc{bytedancese,
  author       = {{bytedance}},
  title        = {{seed1x6}},
  howpublished = {\url{https://seed.bytedance.com/en/seed1x6}},
  year         = {2025}
}

@misc{grok_4,
  author       = {{X.ai}},
  title        = {{grok-4}},
  howpublished = {\url{https://x.ai/news/grok-4}},
  year         = {2025},
  month        = {July},
  day          = {09}
}

@article{zheng2023judging,
  title={Judging llm-as-a-judge with mt-bench and chatbot arena},
  author={Zheng, Lianmin and Chiang, Wei-Lin and Sheng, Ying and Zhuang, Siyuan and Wu, Zhanghao and Zhuang, Yonghao and Lin, Zi and Li, Zhuohan and Li, Dacheng and Xing, Eric and others},
  journal={Advances in neural information processing systems},
  volume={36},
  pages={46595--46623},
  year={2023}
}

@misc{browser-use,
  author       = {{browser-use}},
  title        = {{browser-use}},
  howpublished = {\url{https://browser-use.com/}},
  year         = {2025}
}

@misc{vteam2025glm45vglm41vthinkingversatilemultimodal,
      title={GLM-4.5V and GLM-4.1V-Thinking: Towards Versatile Multimodal Reasoning with Scalable Reinforcement Learning}, 
      author={V Team and Wenyi Hong and Wenmeng Yu and Xiaotao Gu and Guo Wang and Guobing Gan and Haomiao Tang and Jiale Cheng and Ji Qi and Junhui Ji and Lihang Pan and others},
      year={2025},
      eprint={2507.01006},
      archivePrefix={arXiv},
      primaryClass={cs.CV},
      url={https://arxiv.org/abs/2507.01006}, 
}

@misc{kimiteam2025kimivltechnicalreport,
      title={Kimi-VL Technical Report}, 
      author={Kimi Team and Angang Du and Bohong Yin and Bowei Xing and Bowen Qu and Bowen Wang and Cheng Chen and Chenlin Zhang and Chenzhuang Du and Chu Wei and Congcong Wang and others},
      year={2025},
      eprint={2504.07491},
      archivePrefix={arXiv},
      primaryClass={cs.CV},
      url={https://arxiv.org/abs/2504.07491}, 
}

@article{wang2025internvideo2,
  title={Internvideo2. 5: Empowering video mllms with long and rich context modeling},
  author={Wang, Yi and Li, Xinhao and Yan, Ziang and He, Yinan and Yu, Jiashuo and Zeng, Xiangyu and Wang, Chenting and Ma, Changlian and Huang, Haian and Gao, Jianfei and others},
  journal={arXiv preprint arXiv:2501.12386},
  year={2025}
}

@article{openai2025gpt,
  title={Gpt-5 system card},
  author={OpenAI},
  journal={openai.com/index/gpt-5-system-card},
  year={2025},
note={Accessed: 2025-09-04}
}

@article{guo2025seed1,
  title={Seed1. 5-vl technical report},
  author={Guo, Dong and Wu, Faming and Zhu, Feida and Leng, Fuxing and Shi, Guang and Chen, Haobin and Fan, Haoqi and Wang, Jian and Jiang, Jianyu and Wang, Jiawei and others},
  journal={arXiv preprint arXiv:2505.07062},
  year={2025}
}

@article{comanici2025gemini,
  title={Gemini 2.5: Pushing the frontier with advanced reasoning, multimodality, long context, and next generation agentic capabilities},
  author={Comanici, Gheorghe and Bieber, Eric and Schaekermann, Mike and Pasupat, Ice and Sachdeva, Noveen and Dhillon, Inderjit and Blistein, Marcel and Ram, Ori and Zhang, Dan and Rosen, Evan and others},
  journal={arXiv preprint arXiv:2507.06261},
  year={2025}
}

@article{maaz2023video,
  title={Video-chatgpt: Towards detailed video understanding via large vision and language models},
  author={Maaz, Muhammad and Rasheed, Hanoona and Khan, Salman and Khan, Fahad Shahbaz},
  journal={arXiv preprint arXiv:2306.05424},
  year={2023}
}

@misc{gou2025mind2web2evaluatingagentic,
      title={Mind2Web 2: Evaluating Agentic Search with Agent-as-a-Judge}, 
      author={Boyu Gou and Zanming Huang and Yuting Ning and Yu Gu and Michael Lin and Weijian Qi and Andrei Kopanev and Botao Yu and Bernal Jiménez Gutiérrez and Yiheng Shu and Chan Hee Song and Jiaman Wu and Shijie Chen and Hanane Nour Moussa and Tianshu Zhang and Jian Xie and Yifei Li and Tianci Xue and Zeyi Liao and Kai Zhang and Boyuan Zheng and Zhaowei Cai and Viktor Rozgic and Morteza Ziyadi and Huan Sun and Yu Su},
      year={2025},
      eprint={2506.21506},
      archivePrefix={arXiv},
      primaryClass={cs.AI},
      url={https://arxiv.org/abs/2506.21506}, 
}

@inproceedings{jimenez2024swe,
  title={SWE-bench: Can Language Models Resolve Real-world Github Issues?},
  author={Jimenez, Carlos E and Yang, John and Wettig, Alexander and Yao, Shunyu and Pei, Kexin and Press, Ofir and Narasimhan, Karthik R},
  booktitle={ICLR},
  year={2024}
}

@inproceedings{yangswe,
  title={SWE-bench Multimodal: Do AI Systems Generalize to Visual Software Domains?},
  author={Yang, John and Jimenez, Carlos E and Zhang, Alex L and Lieret, Kilian and Yang, Joyce and Wu, Xindi and Press, Ori and Muennighoff, Niklas and Synnaeve, Gabriel and Narasimhan, Karthik R and others},
  booktitle={The Thirteenth International Conference on Learning Representations},
  year={2024}
}

@inproceedings{mialon2023gaia,
  title={Gaia: a benchmark for general ai assistants},
  author={Mialon, Gr{\'e}goire and Fourrier, Cl{\'e}mentine and Wolf, Thomas and LeCun, Yann and Scialom, Thomas},
  booktitle={The Twelfth International Conference on Learning Representations},
  year={2023}
}

@inproceedings{lu23mathvista,
  title={MathVista: Evaluating Mathematical Reasoning of Foundation Models in Visual Contexts},
  author={Lu, Pan and Bansal, Hritik and Xia, Tony and Liu, Jiacheng and Li, Chunyuan and Hajishirzi, Hannaneh and Cheng, Hao and Chang, Kai-Wei and Galley, Michel and Gao, Jianfeng},
  booktitle={The 3rd Workshop on Mathematical Reasoning and AI at NeurIPS'23},
  year={2023}
}

@misc{li2023seedbench2,
      title={SEED-Bench-2: Benchmarking Multimodal Large Language Models}, 
      author={Bohao Li and Yuying Ge and Yixiao Ge and Guangzhi Wang and Rui Wang and Ruimao Zhang and Ying Shan},
      year={2023},
      eprint={2311.17092},
      archivePrefix={arXiv},
      primaryClass={cs.CV},
      url={https://arxiv.org/abs/2311.17092}, 
}

@article{wang2024muirbench,
  title={Muirbench: A comprehensive benchmark for robust multi-image understanding},
  author={Wang, Fei and Fu, Xingyu and Huang, James Y and Li, Zekun and Liu, Qin and Liu, Xiaogeng and Ma, Mingyu Derek and Xu, Nan and Zhou, Wenxuan and Zhang, Kai and others},
  journal={arXiv preprint arXiv:2406.09411},
  year={2024}
}

@article{hu2025video,
  title={Video-mmmu: Evaluating knowledge acquisition from multi-discipline professional videos},
  author={Hu, Kairui and Wu, Penghao and Pu, Fanyi and Xiao, Wang and Zhang, Yuanhan and Yue, Xiang and Li, Bo and Liu, Ziwei},
  journal={arXiv preprint arXiv:2501.13826},
  year={2025}
}

@article{zhuge2024agent,
  title={Agent-as-a-judge: Evaluate agents with agents},
  author={Zhuge, Mingchen and Zhao, Changsheng and Ashley, Dylan and Wang, Wenyi and Khizbullin, Dmitrii and Xiong, Yunyang and Liu, Zechun and Chang, Ernie and Krishnamoorthi, Raghuraman and Tian, Yuandong and others},
  journal={arXiv preprint arXiv:2410.10934},
  year={2024}
}

@inproceedings{li2024mvbench,
  title={Mvbench: A comprehensive multi-modal video understanding benchmark},
  author={Li, Kunchang and Wang, Yali and He, Yinan and Li, Yizhuo and Wang, Yi and Liu, Yi and Wang, Zun and Xu, Jilan and Chen, Guo and Luo, Ping and others},
  booktitle={Proceedings of the IEEE/CVF Conference on Computer Vision and Pattern Recognition},
  pages={22195--22206},
  year={2024}
}

@article{fang2024mmbench,
  title={Mmbench-video: A long-form multi-shot benchmark for holistic video understanding},
  author={Fang, Xinyu and Mao, Kangrui and Duan, Haodong and Zhao, Xiangyu and Li, Yining and Lin, Dahua and Chen, Kai},
  journal={Advances in Neural Information Processing Systems},
  volume={37},
  pages={89098--89124},
  year={2024}
}

@inproceedings{fu2025video,
  title={Video-mme: The first-ever comprehensive evaluation benchmark of multi-modal llms in video analysis},
  author={Fu, Chaoyou and Dai, Yuhan and Luo, Yongdong and Li, Lei and Ren, Shuhuai and Zhang, Renrui and Wang, Zihan and Zhou, Chenyu and Shen, Yunhang and Zhang, Mengdan and others},
  booktitle={Proceedings of the Computer Vision and Pattern Recognition Conference},
  pages={24108--24118},
  year={2025}
}

@article{si2024design2code,
  title={Design2code: Benchmarking multimodal code generation for automated front-end engineering},
  author={Si, Chenglei and Zhang, Yanzhe and Li, Ryan and Yang, Zhengyuan and Liu, Ruibo and Yang, Diyi},
  journal={arXiv preprint arXiv:2403.03163},
  year={2024}
}

@inproceedings{yue2024mmmu,
  title={Mmmu: A massive multi-discipline multimodal understanding and reasoning benchmark for expert agi},
  author={Yue, Xiang and Ni, Yuansheng and Zhang, Kai and Zheng, Tianyu and Liu, Ruoqi and Zhang, Ge and Stevens, Samuel and Jiang, Dongfu and Ren, Weiming and Sun, Yuxuan and others},
  booktitle={Proceedings of the IEEE/CVF Conference on Computer Vision and Pattern Recognition},
  pages={9556--9567},
  year={2024}
}

@article{awal2025webmmu,
  title={WebMMU: A benchmark for multimodal multilingual website understanding and code generation},
  author={Awal, Rabiul and Massoud, Mahsa and Feizi, Aarash and Li, Zichao and Wang, Suyuchen and Pal, Christopher and Agrawal, Aishwarya and Vazquez, David and Reddy, Siva and Rodriguez, Juan A and others},
  journal={arXiv preprint arXiv:2508.16763},
  year={2025}
}

@article{xu2025web,
  title={Web-bench: A llm code benchmark based on web standards and frameworks},
  author={Xu, Kai and Mao, YiWei and Guan, XinYi and Feng, ZiLong},
  journal={arXiv preprint arXiv:2505.07473},
  year={2025}
}

@article{ning2023video,
  title={Video-bench: A comprehensive benchmark and toolkit for evaluating video-based large language models},
  author={Ning, Munan and Zhu, Bin and Xie, Yujia and Lin, Bin and Cui, Jiaxi and Yuan, Lu and Chen, Dongdong and Yuan, Li},
  journal={arXiv preprint arXiv:2311.16103},
  year={2023}
}

@inproceedings{chen2024autoeval,
  title={Autoeval-video: An automatic benchmark for assessing large vision language models in open-ended video question answering},
  author={Chen, Xiuyuan and Lin, Yuan and Zhang, Yuchen and Huang, Weiran},
  booktitle={European Conference on Computer Vision},
  pages={179--195},
  year={2024},
  organization={Springer}
}

@inproceedings{liu2024mmbench,
  title={Mmbench: Is your multi-modal model an all-around player?},
  author={Liu, Yuan and Duan, Haodong and Zhang, Yuanhan and Li, Bo and Zhang, Songyang and Zhao, Wangbo and Yuan, Yike and Wang, Jiaqi and He, Conghui and Liu, Ziwei and others},
  booktitle={European conference on computer vision},
  pages={216--233},
  year={2024},
  organization={Springer}
}

@misc{lu2025webgenbenchevaluatingllmsgenerating,
      title={WebGen-Bench: Evaluating LLMs on Generating Interactive and Functional Websites from Scratch}, 
      author={Zimu Lu and Yunqiao Yang and Houxing Ren and Haotian Hou and Han Xiao and Ke Wang and Weikang Shi and Aojun Zhou and Mingjie Zhan and Hongsheng Li},
      year={2025},
      eprint={2505.03733},
      archivePrefix={arXiv},
      primaryClass={cs.CL},
      url={https://arxiv.org/abs/2505.03733}, 
}

@article{hurst2024gpt,
  title={Gpt-4o system card},
  author={Hurst, Aaron and Lerer, Adam and Goucher, Adam P and Perelman, Adam and Ramesh, Aditya and Clark, Aidan and Ostrow, AJ and Welihinda, Akila and Hayes, Alan and Radford, Alec and others},
  journal={arXiv preprint arXiv:2410.21276},
  year={2024}
}

@article{guo2024iw,
  title={Iw-bench: Evaluating large multimodal models for converting image-to-web},
  author={Guo, Hongcheng and Zhang, Wei and Chen, Junhao and Gu, Yaonan and Yang, Jian and Du, Junjia and Hui, Binyuan and Liu, Tianyu and Ma, Jianxin and Zhou, Chang and others},
  journal={arXiv preprint arXiv:2409.18980},
  year={2024}
}

@inproceedings{caron2021emerging,
  title={Emerging properties in self-supervised vision transformers},
  author={Caron, Mathilde and Touvron, Hugo and Misra, Ishan and J{\'e}gou, Herv{\'e} and Mairal, Julien and Bojanowski, Piotr and Joulin, Armand},
  booktitle={Proceedings of the IEEE/CVF international conference on computer vision},
  pages={9650--9660},
  year={2021}
}

@article{oquab2023dinov2,
  title={Dinov2: Learning robust visual features without supervision},
  author={Oquab, Maxime and Darcet, Timoth{\'e}e and Moutakanni, Th{\'e}o and Vo, Huy and Szafraniec, Marc and Khalidov, Vasil and Fernandez, Pierre and Haziza, Daniel and Massa, Francisco and El-Nouby, Alaaeldin and others},
  journal={arXiv preprint arXiv:2304.07193},
  year={2023}
}

@inproceedings{zhang2018unreasonable,
  title={The unreasonable effectiveness of deep features as a perceptual metric},
  author={Zhang, Richard and Isola, Phillip and Efros, Alexei A and Shechtman, Eli and Wang, Oliver},
  booktitle={Proceedings of the IEEE conference on computer vision and pattern recognition},
  pages={586--595},
  year={2018}
}

@inproceedings{radford2021learning,
  title={Learning transferable visual models from natural language supervision},
  author={Radford, Alec and Kim, Jong Wook and Hallacy, Chris and Ramesh, Aditya and Goh, Gabriel and Agarwal, Sandhini and Sastry, Girish and Askell, Amanda and Mishkin, Pamela and Clark, Jack and others},
  booktitle={International conference on machine learning},
  pages={8748--8763},
  year={2021},
  organization={PmLR}
}

@misc{xiao2025interaction2code,
      title={Interaction2Code: Benchmarking MLLM-based Interactive Webpage Code Generation from Interactive Prototyping}, 
      author={Jingyu Xiao and Yuxuan Wan and Yintong Huo and Zixin Wang and Xinyi Xu and Wenxuan Wang and Zhiyao Xu and Yuhang Wang and Michael R. Lyu},
      year={2025},
      eprint={2411.03292},
      archivePrefix={arXiv},
      primaryClass={cs.SE},
      url={https://arxiv.org/abs/2411.03292}, 
}

@inproceedings{luo2024layoutllm,
  title={Layoutllm: Layout instruction tuning with large language models for document understanding},
  author={Luo, Chuwei and Shen, Yufan and Zhu, Zhaoqing and Zheng, Qi and Yu, Zhi and Yao, Cong},
  booktitle={Proceedings of the IEEE/CVF conference on computer vision and pattern recognition},
  pages={15630--15640},
  year={2024}
}

@inproceedings{gupta2023visual,
  title={Visual programming: Compositional visual reasoning without training},
  author={Gupta, Tanmay and Kembhavi, Aniruddha},
  booktitle={Proceedings of the IEEE/CVF conference on computer vision and pattern recognition},
  pages={14953--14962},
  year={2023}
}

@article{song2025tianyue,
  title={VisualPuzzles: Decoupling Multimodal Reasoning Evaluation from Domain Knowledge},
  author={Song, Yueqi and Ou, Tianyue and Kong, Yibo and Li, Zecheng and Neubig, Graham and Yue, Xiang},
  journal={arXiv preprint arXiv:2504.10342},
  year={2025}
}

@article{cui2025paddleocr,
  title={Paddleocr 3.0 technical report},
  author={Cui, Cheng and Sun, Ting and Lin, Manhui and Gao, Tingquan and Zhang, Yubo and Liu, Jiaxuan and Wang, Xueqing and Zhang, Zelun and Zhou, Changda and Liu, Hongen and others},
  journal={arXiv preprint arXiv:2507.05595},
  year={2025}
}
\bibliographystyle{iwr}

\newpage
\appendix
\lstset{
    basicstyle=\sffamily\small, 
    keywordstyle=\color{blue}\bfseries,   
    commentstyle=\color{gray},         
    stringstyle=\color{purple},          
    breaklines=true,            
    breakatwhitespace=true,     
    postbreak=\mbox{\textcolor{red}{$\hookrightarrow$}\space}, 
    showstringspaces=false,     
    frame=none,                 
    tabsize=2,                  
    captionpos=b,               
}

\section{Multi-Dimensional Task Taxonomy}
\label{sec:app_tasktax}

To move beyond a monolithic view of difficulty and enable a fine-grained analysis of model capabilities, we developed a three-dimensional taxonomy to classify each task. This taxonomy categorizes tasks along the orthogonal axes of Interaction Complexity, Visual Complexity, and Application Domain, providing a structured framework to understand the specific challenges inherent in each task and to diagnose model failure modes with high precision.

\paragraph{Interaction Complexity (L1-L4)} This first axis categorizes tasks based on the depth of logical and temporal understanding required for successful reconstruction.
\begin{itemize}
    \item \textbf{L1: Static Content Consumption.} Tasks involve passive information consumption, primarily requiring correct handling of vertical scrolling to reconstruct long pages that extend beyond a single viewport (e.g., browsing a blog post or a project's README).
    \item \textbf{L2: Simple State Manipulation.} Tasks feature components that manage local state, such as filtering e-commerce results, switching between on-page tabs, or expanding/collapsing accordion menus. This level tests the generation of basic client-side event handlers.
    \item \textbf{L3: Complex Workflow Interaction.} These tasks involve multi-step, sequential interactions where state is passed between components, such as a multi-step product configurator or an online booking process. This tests understanding of application logic and inter-component communication.
    \item \textbf{L4: Algorithmic/Game Logic.} The most complex level requires the model to reverse-engineer and implement a set of rules or algorithms, such as an online calculator, a text-based puzzle, or a simple game like 2048.
\end{itemize}

\paragraph{Visual Complexity (V1-V4)} This second axis captures the static challenge of rendering the webpage's appearance, focusing on its layout and styling.
\begin{itemize}
    \item \textbf{V1: Minimalist Layouts.} Simple, single- or two-column structures with standard element alignment, typical of documentation or text-heavy sites.
    \item \textbf{V2: Standard Grid-based Layouts.} Organized grid systems are used in e-commerce or news portals, featuring numerous but regularly arranged elements.
    \item \textbf{V3: Asymmetric \& Modern Layouts.} Visually-driven designs with complex CSS, such as overlapping elements, parallax scrolling, and non-standard component shapes.
    \item \textbf{V4: Data-Dense Layouts.} Dashboards or admin panels with a high density of information presented in charts, tables, and data cards, testing the ability to generate precise, repetitive structures.
\end{itemize}

\paragraph{Application Domain} To ensure our benchmark reflects the breadth of real-world web applications, the third axis classifies tasks by their Application Domain. Drawing inspiration from established taxonomies in web-centric agent research \citep{gou2025mind2web2evaluatingagentic}, we group tasks into five high-level domains: \textbf{Commerce \& Services} (e-commerce, booking, finance), \textbf{Knowledge \& Education} (academic sites, news portals, documentation), \textbf{Productivity \& Tools} (calculators, project management boards), \textbf{Entertainment \& Media} (games, streaming platforms), and \textbf{Lifestyle \& Community} (social forums, blogs). This classification guarantees that models are evaluated across a diverse spectrum of functionalities and visual paradigms. For instance, a task involving filtering products on an e-commerce site is tagged as [L2, V2, Commerce], while a task requiring the reconstruction of a simple browser game is tagged as [L4, V1, Entertainment]. This multi-dimensional labeling allows us to analyze whether a model's performance correlates with certain types of interactions, visual styles, or application contexts.

\section{Detailed Experimental Results}
\label{appendix:detailed_results}

This appendix provides a comprehensive breakdown of model performance on IWR-Bench across the three classification axes defined in the taxonomy: Application Domain (Table~\ref{tab:domain_breakdown_final}), Interaction Logic Complexity (Table~\ref{tab:interaction_breakdown_final}), and Visual Complexity (Table~\ref{tab:visual_breakdown_final}). For each task category, the Final Score is reported. 

\definecolor{Gray}{gray}{0.9}

\begin{table*}[!ht]
\centering
\small
\caption{Final Score breakdown by Interaction Logic Complexity. }
\label{tab:interaction_breakdown_final}
\resizebox{\textwidth}{!}{%
\begin{tabular}{lcccc}
\toprule
\textbf{Model} & \thead{Static Content \\ Consumption (L1)} & \thead{Simple State \\ Manipulation (L2)} & \thead{Complex Workflow \\ Interaction (L3)} & \thead{Algorithmic/Game \\ Logic (L4)} \\
\midrule
\multicolumn{5}{l}{\textit{Proprietary MLLMs}} \\
\quad GPT-5 & 61.85 & 35.43 & 35.12 & 25.26 \\
\rowcolor{Gray}
\quad Claude-Sonnet-4 (thinking) & 65.75 & 33.88 & 31.05 & 25.86 \\
\rowcolor{Gray}
\quad Claude-Opus-4 (thinking) & 61.78 & 31.52 & 32.57 & 30.07 \\
\quad Doubao-seed-1.6 & 63.69 & 34.03 & 30.12 & 24.08 \\
\quad Claude-Sonnet-4 & 68.36 & 31.04 & 32.86 & 25.15 \\
\quad Claude-Opus-4 & 66.88 & 30.85 & 31.30 & 25.76 \\
\quad GPT-5-mini & 56.83 & 32.10 & 31.38 & 26.84 \\
\quad GPT-4.1 & 51.45 & 30.32 & 31.98 & 24.78 \\
\rowcolor{Gray}
\quad Gemini-2.5-Pro (thinking) & 68.96 & 26.88 & 28.15 & 23.83 \\
\quad Gemini-2.5-Pro & 59.58 & 31.51 & 23.95 & 22.58 \\
\quad GPT-4o (latest) & 48.61 & 29.09 & 27.53 & 23.75 \\
\quad Gemini-2.5-Flash & 56.57 & 24.95 & 25.53 & 15.00 \\
\quad GPT-5-nano & 46.23 & 25.57 & 24.31 & 19.02 \\
\quad Grok-4 & 48.37 & 24.94 & 22.76 & 19.53 \\
\quad GPT-4o (0806) & 38.59 & 24.95 & 24.65 & 17.90 \\
\quad Doubao-seed-1.6-flash & 42.77 & 22.14 & 22.65 & 13.88 \\
\quad Gemini-2.5-Flash-Lite & 35.34 & 15.61 & 13.87 & 14.70 \\
\midrule
\multicolumn{5}{l}{\textit{Open-Source MLLMs}} \\
\rowcolor{Gray}
\quad Qwen3-VL (thinking) & 51.05 & 30.43 & 29.86 & 23.60 \\
\quad Qwen2.5-VL-72B & 45.35 & 22.83 & 21.90 & 15.46 \\
\quad Qwen2.5-VL-32B & 37.64 & 20.45 & 18.80 & 17.48 \\
\quad Keye-VL-1.5-8B & 46.85 & 19.47 & 12.11 & 10.13 \\
\quad MiniCPM-V-4.5 & 37.30 & 18.02 & 15.27 & 11.10 \\
\quad Qwen2.5-VL-7B & 28.58 & 15.99 & 13.91 & 8.21 \\
\rowcolor{Gray}
\quad Kimi-VL (thinking) & 23.61 & 15.45 & 10.63 & 12.75 \\
\quad Mimo-VL-7B & 22.23 & 12.32 & 9.36 & 7.87 \\
\quad GLM-4.5V & 18.56 & 11.02 & 9.77 & 10.08 \\
\midrule
\multicolumn{5}{l}{\textit{Open-Source Video-Specialized LMs}} \\
\quad VideoLLaMA3-7B & 23.11 & 14.15 & 12.31 & 8.77 \\
\quad InternVideo-2.5-Chat-8B & 27.89 & 9.81 & 7.68 & 5.42 \\
\bottomrule
\end{tabular}}
\end{table*}

\begin{table*}[!ht]
\centering
\small
\caption{Final Score breakdown by Visual Complexity.}
\label{tab:visual_breakdown_final}
\resizebox{\textwidth}{!}{%
\begin{tabular}{lcccc}
\toprule
\textbf{Model} & \thead{Minimalist \\ Layouts (V1)} & \thead{Standard \\ Grid-based Layouts (V2)} & \thead{Asymmetric \& \\ Modern Layouts (V3)} & \thead{Data-Dense \\ Layouts (V4)} \\
\midrule
\multicolumn{5}{l}{\textit{Proprietary MLLMs}} \\
\quad GPT-5 & 44.77 & 30.73 & 43.77 & 26.05 \\
\rowcolor{Gray}
\quad Claude-Sonnet-4 (thinking) & 40.05 & 31.12 & 40.08 & 25.26 \\
\rowcolor{Gray}
\quad Claude-Opus-4 (thinking) & 37.01 & 32.26 & 38.95 & 22.97 \\
\quad Doubao-seed-1.6 & 41.55 & 30.56 & 38.03 & 26.97 \\
\quad Claude-Sonnet-4 & 37.33 & 30.28 & 39.95 & 26.25 \\
\quad Claude-Opus-4 & 35.77 & 30.79 & 38.34 & 24.77 \\
\quad GPT-5-mini & 38.76 & 29.29 & 38.44 & 26.75 \\
\quad GPT-4.1 & 35.24 & 28.96 & 36.89 & 25.32 \\
\rowcolor{Gray}
\quad Gemini-2.5-Pro (thinking) & 31.59 & 25.77 & 36.99 & 25.82 \\
\quad Gemini-2.5-Pro & 37.20 & 25.17 & 36.51 & 23.27 \\
\quad GPT-4o (latest) & 31.89 & 26.14 & 34.64 & 24.41 \\
\quad Gemini-2.5-Flash & 32.21 & 21.80 & 34.13 & 16.28 \\
\quad GPT-5-nano & 29.21 & 24.39 & 28.96 & 20.37 \\
\quad Grok-4 & 33.73 & 22.11 & 30.01 & 17.18 \\
\quad GPT-4o (0806) & 29.88 & 23.54 & 27.41 & 19.62 \\
\quad Doubao-seed-1.6-flash & 30.23 & 22.76 & 22.87 & 17.27 \\
\quad Gemini-2.5-Flash-Lite & 19.64 & 15.44 & 19.31 & 8.31 \\
\midrule
\multicolumn{5}{l}{\textit{Open-Source MLLMs}} \\
\rowcolor{Gray}
\quad Qwen3-VL (thinking) & 38.14 & 28.74 & 33.37 & 26.25 \\
\quad Qwen2.5-VL-72B & 28.68 & 21.14 & 28.31 & 12.48 \\
\quad Qwen2.5-VL-32B & 27.56 & 17.90 & 24.06 & 16.15 \\
\quad Keye-VL-1.5-8B & 24.75 & 15.65 & 20.85 & 12.74 \\
\quad MiniCPM-V-4.5 & 28.24 & 16.11 & 18.87 & 10.79 \\
\quad Qwen2.5-VL-7B & 20.48 & 12.68 & 18.52 & 11.48 \\
\rowcolor{Gray}
\quad Kimi-VL (thinking) & 18.57 & 14.22 & 14.21 & 9.26 \\
\quad Mimo-VL-7B & 14.68 & 11.42 & 12.42 & 6.71 \\
\quad GLM-4.5V & 12.39 & 10.66 & 13.34 & 4.20 \\
\midrule
\multicolumn{5}{l}{\textit{Open-Source Video-Specialized LMs}} \\
\quad VideoLLaMA3-7B & 18.81 & 13.55 & 13.69 & 8.56 \\
\quad InternVideo-2.5-Chat-8B & 13.36 & 11.07 & 9.09 & 5.57 \\
\bottomrule
\end{tabular}
}
\end{table*}

\begin{table*}[!ht]
\centering
\small
\caption{Final Score breakdown by Application Domain.}
\label{tab:domain_breakdown_final}
\resizebox{\textwidth}{!}{%
\begin{tabular}{lccccc}
\toprule
\textbf{Model} & \thead{Business \& \\ Services} & \thead{Entertainment \\ \& Media} & \thead{Knowledge \& \\ Education} & \thead{Life \& \\ Community} & \thead{Productivity \\ \& Tools} \\
\midrule
\multicolumn{6}{l}{\textit{Proprietary MLLMs}} \\
\quad GPT-5 & 39.37 & 47.24 & 37.05 & 22.80 & 28.53 \\
\rowcolor{Gray}
\quad Claude-Sonnet-4 (thinking) & 39.74 & 41.02 & 32.60 & 24.86 & 31.15 \\
\rowcolor{Gray}
\quad Claude-Opus-4 (thinking) & 35.54 & 42.48 & 34.91 & 24.21 & 28.56 \\
\quad Doubao-seed-1.6 & 37.82 & 43.19 & 32.14 & 22.44 & 30.20 \\
\quad Claude-Sonnet-4 & 38.84 & 42.09 & 32.22 & 25.57 & 27.58 \\
\quad Claude-Opus-4 & 38.64 & 43.25 & 30.51 & 22.03 & 28.09 \\
\quad GPT-5-mini & 37.16 & 45.11 & 30.29 & 21.38 & 28.36 \\
\quad GPT-4.1 & 33.58 & 45.35 & 27.61 & 17.28 & 32.65 \\
\rowcolor{Gray}
\quad Gemini-2.5-Pro (thinking) & 31.49 & 41.67 & 30.31 & 15.82 & 26.34 \\
\quad Gemini-2.5-Pro & 33.66 & 36.93 & 30.74 & 19.03 & 25.55 \\
\quad GPT-4o (latest) & 36.33 & 32.35 & 28.97 & 20.01 & 25.42 \\
\quad Gemini-2.5-Flash & 29.16 & 40.60 & 21.38 & 15.27 & 25.83 \\
\quad GPT-5-nano & 30.61 & 34.29 & 22.67 & 16.50 & 23.69 \\
\quad Grok-4 & 27.71 & 32.06 & 26.47 & 15.00 & 19.85 \\
\quad GPT-4o (0806) & 27.60 & 32.05 & 23.97 & 15.54 & 23.29 \\
\quad Doubao-seed-1.6-flash & 25.37 & 31.07 & 20.17 & 14.21 & 22.36 \\
\quad Gemini-2.5-Flash-Lite & 18.38 & 24.93 & 10.13 & 9.42 & 20.55 \\
\midrule
\multicolumn{6}{l}{\textit{Open-Source MLLMs}} \\
\rowcolor{Gray}
\quad Qwen3-VL (thinking) & 37.36 & 35.00 & 29.57 & 17.71 & 31.19 \\
\quad Qwen2.5-VL-72B & 28.71 & 31.20 & 20.89 & 14.29 & 19.88 \\
\quad Qwen2.5-VL-32B & 24.05 & 28.58 & 20.67 & 9.80 & 16.73 \\
\quad Keye-VL-1.5-8B & 19.27 & 26.29 & 16.42 & 9.95 & 16.50 \\
\quad MiniCPM-V-4.5 & 17.10 & 26.63 & 17.38 & 11.21 & 14.63 \\
\quad Qwen2.5-VL-7B & 17.11 & 21.96 & 16.69 & 4.67 & 11.61 \\
\rowcolor{Gray}
\quad Kimi-VL (thinking) & 16.77 & 17.58 & 13.87 & 9.59 & 10.83 \\
\quad Mimo-VL-7B & 12.90 & 14.64 & 12.11 & 4.33 & 11.08 \\
\quad GLM-4.5V & 10.78 & 18.20 & 10.10 & 2.88 & 11.10 \\
\midrule
\multicolumn{6}{l}{\textit{Open-Source Video-Specialized LMs}} \\
\quad VideoLLaMA3-7B & 14.36 & 16.44 & 14.44 & 9.21 & 11.51 \\
\quad InternVideo-2.5-Chat-8B & 10.77 & 15.77 & 9.18 & 2.90 & 9.36 \\
\bottomrule
\end{tabular}
}
\end{table*}

\section{Metric Parameter Validation}
\label{sec:app_metric_par}
The VFS and Final Score metrics rely on the weighting coefficients $w$ and $\alpha$. These parameters are determined and validated through a human alignment study. A sample of 60 evaluation instances is constructed by selecting outputs from three randomly chosen models for each of 20 randomly selected tasks from IWR-Bench. Each instance is assessed by five PhD-level students on two dimensions: visual fidelity and overall quality.
To determine the optimal parameters, a grid search is performed over the discrete set $\{0.1, 0.2, \dots, 0.9\}$ for both $w$ and $\alpha$. The value that maximizes the Spearman's $\rho$ correlation between the automated scores and the aggregated human judgments is selected. For the VFS metric, the peak correlation ($\rho=0.57$) is observed at $w=0.5$, indicating an equal weighting between LVS and HVS. For the Final Score, the maximum correlation with human overall judgment ($\rho=0.65$) is achieved with $\alpha=0.7$, empirically validating the decision to weigh functionality (IFS) more heavily than visual fidelity (VFS).

\section{Task and Action Representation}
\label{sec:app_task_action}
Each task in IWR-Bench is formally defined by an `action\_sequence`, a structured list of discrete actions that an automated agent must perform to validate the reconstructed webpage. This representation standardizes the evaluation process. We defined a vocabulary of atomic actions, including \textbf{Click(description)}, \textbf{Type(key, description)}, \textbf{Scroll(direction, amount, description)}, and \textbf{Press(key, description)}. A crucial design choice is the use of a natural language `description` field for targeting elements instead of unstable positional coordinates (e.g., "Click the primary 'Submit' button" instead of "Click at (x:120, y:350)"). This makes the evaluation robust to minor layout variations in the generated code and tests a more semantic understanding of the page structure, both for the model during generation and the agent during evaluation.

\section{Prompts}
\label{sec:app_prompts}

A standardized system prompt is employed for all models and tasks in IWR-Bench to ensure a fair evaluation. This prompt defines clear requirements for the task, output format, and operational constraints. Such a design minimizes ambiguity and helps isolate the core code generation capabilities of each model. The complete prompt template is detailed in Figure~\ref{fig:prompt_chart_system}.

\begin{figure*}[!ht]
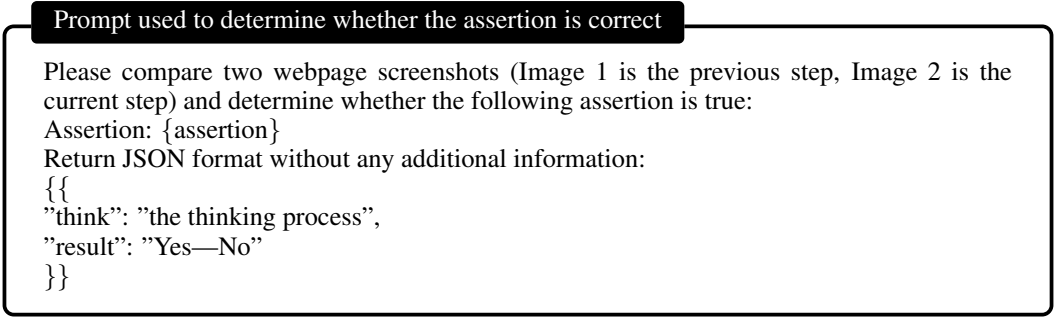

\begin{AIbox}{Prompt used to determine whether the assertion is correct}
    {
Please compare two webpage screenshots (Image 1 is the previous step, Image 2 is the current step) and determine whether the following assertion is true:

Assertion: \{assertion\}

Return JSON format without any additional information:

\{\{

    "think": "the thinking process",
    
    "result": "Yes|No"
    
\}\}

}
\end{AIbox}
\caption{Prompt used to determine whether the assertion is correct}
\label{fig:prompt_assertion}
\end{figure*}

The evaluation relies on a large multimodal model guided by two distinct prompts. To assess the similarity between generated and reference webpages, a prompt template is utilized (Figure~\ref{fig:prompt_hvs}). This template instructs the model to perform both quantitative and qualitative evaluations. For logical assertion verification, a separate prompt, presented in Figure~\ref{fig:prompt_assertion}, is employed to determine the correctness of an action.

\begin{figure*}[!ht]
\begin{AIbox}{Prompt template for the IWR-Bench}
    {
You are an expert front-end developer. Your task is to create a pixel-perfect replica of a website from a video.

Generate a single `index.html' file that contains all HTML, CSS, and JavaScript necessary to replicate the UI, content, and interaction features shown. The webpage resolution in the video is \textcolor{brown}{\textless resolution\textgreater}.

Instructions:

1. Single File Output: All HTML, CSS, and JS must be in one `index.html' file.

2. If backend logic is implied, mock it in JS with static data (e.g., a JS array for a fake API call).

3. Assets(Images and Videos in the webpage):

\quad $-$ All images must use the provided stitched image assets.

\quad $-$ The `src' attribute must start with the literal, unchanging string `$\_\_$PLACEHOLDER$\_$ASSETS$\_$BASE$\_$DIR$\_\_/$', followed by the actual filename identified from the stitched image.

\quad $-$ For example: `src=``$\_\_$PLACEHOLDER$\_$ASSETS$\_$BASE$\_$DIR$\_\_/$asset001.svg"'.

\quad $-$ `$<$img$>$' tags must include `width' and `height' attributes.

\quad $-$ The provided stitched image assets are before the video.

4. No External Dependencies: The generated code must be entirely self-contained. No External Libraries and no External Fonts.

5. Final Response: Return \textbf{only the complete HTML code} in a single ```html code block, with no additional text or explanations.

}
\end{AIbox}
\caption{Prompt template for the IWR-Bench}
\label{fig:prompt_chart_system}
\end{figure*}

\begin{figure*}[!ht]
\begin{AIbox}{Prompt for evaluating HVS}
    {
You are an expert Webpage Evaluator. Your task is to provide a quantitative and qualitative assessment of the similarity between a generated webpage and a reference webpage.

The default score is 0.

Evaluation Format:

---

Comments:

-Layout (10 points): \$\{comment and subscore\}

-Elements (15 points): \$\{comment and subscore\}

-Content and Text (40 points): \$\{comment and subscore\}

-Style (15 points): \$\{comment and subscore\}

-Overall (20 points): \$\{comment and subscore\}

Score: \$\{final score\}/100

}
\end{AIbox}
\caption{Prompt for evaluating HVS}
\label{fig:prompt_hvs}
\end{figure*}

\section{Case Study}
\label{sec:app_case_study}
This section presents a selection of representative tasks from the IWR-Bench to illustrate the diversity of challenges encompassed by our benchmark. The input of each case includes a webpage operation video and the static resources involved in the webpage. Then the web pages generated by different multimodal large models and the corresponding interaction results are displayed. Then we provide a detailed analysis of these representative cases, corresponding to the figures presented below. Each analysis breaks down the task's objectives, its position within our taxonomy, and the specific model behaviors observed, illustrating how our benchmark facilitates a fine-grained diagnosis of model capabilities.

\paragraph{Case 1 Analysis: E-commerce Workflow Simulation.}
Our first case study, classified as [L2, V2, E-commerce], simulates a fundamental e-commerce user journey to test a model's ability to handle sequential state manipulations within a standard visual structure. The task requires the model to replicate a workflow involving  filtering a product grid by a specific brand, sorting the filtered results by price and adding a selected item to the shopping cart.

\begin{figure}[h]
    \centering
    \includegraphics[width=0.92\linewidth]{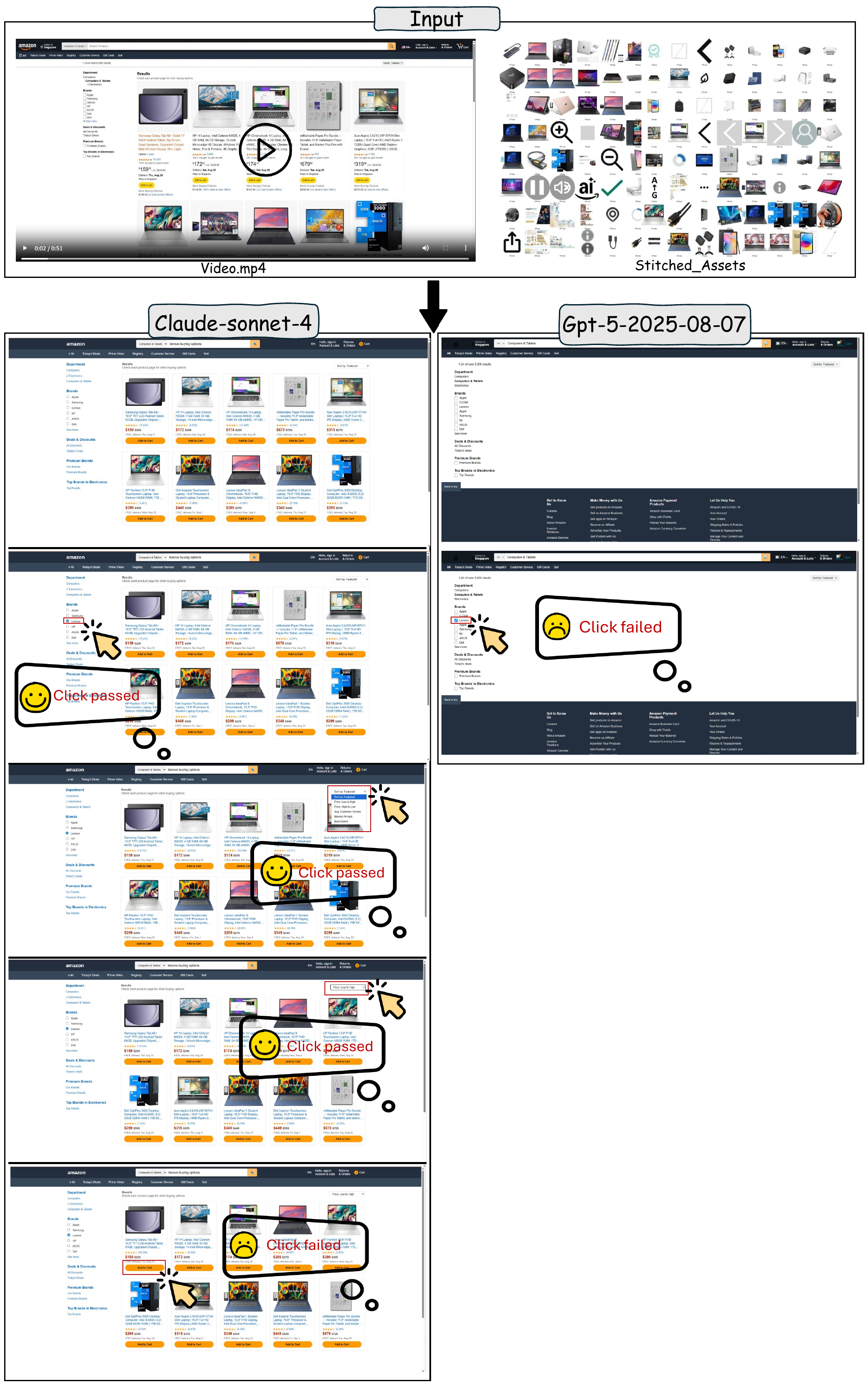}
    \caption{
        \textbf{Case 1: Multi-Step E-commerce Workflow.} This task, classified as \textbf{[L2, V2, E-commerce]}, requires reconstructing a core e-commerce workflow involving filtering products, sorting the results, and adding an item to the shopping cart. 
    }
    \label{fig:case1}
\end{figure}

As illustrated in Figure \ref{fig:case1}, this task effectively exposes different failure modes in different models. On the left, Claude-Sonnet-4 demonstrates good capabilities in static replication and simple state management. It accurately renders the initial layout and correctly implements the action for filtering and sorting. However, its failure occurs at the final "add to cart" step. The right side shows the result of GPT-5. This case likely involved too many static resources, causing the product list on the initial page to fail to render successfully. By pinpointing these different stages of failure, the benchmark provides a granular diagnosis of each model's specific strengths and weaknesses in front-end code generation.

\paragraph{Case 2 Analysis: Algorithmic Logic Reconstruction.}
This case study moves to the highest level of our interaction complexity scale, L4, to assess a model's capacity for algorithmic reasoning. Classified as [L4, V2, Gaming], the task requires the model to reverse-engineer and implement the complete set of rules for a simple browser game (e.g., 2048) based solely on observing its behavior in the input video. The visual complexity is simple (V2), deliberately shifting the evaluation focus from layout replication to the correctness of the underlying algorithmic logic. The The core challenge is to deduce and codify the game's state-transition functions, including tile movement, merging logic, and the spawning of new tiles.

\begin{figure}[h]
    \centering
    \includegraphics[width=0.92\linewidth]{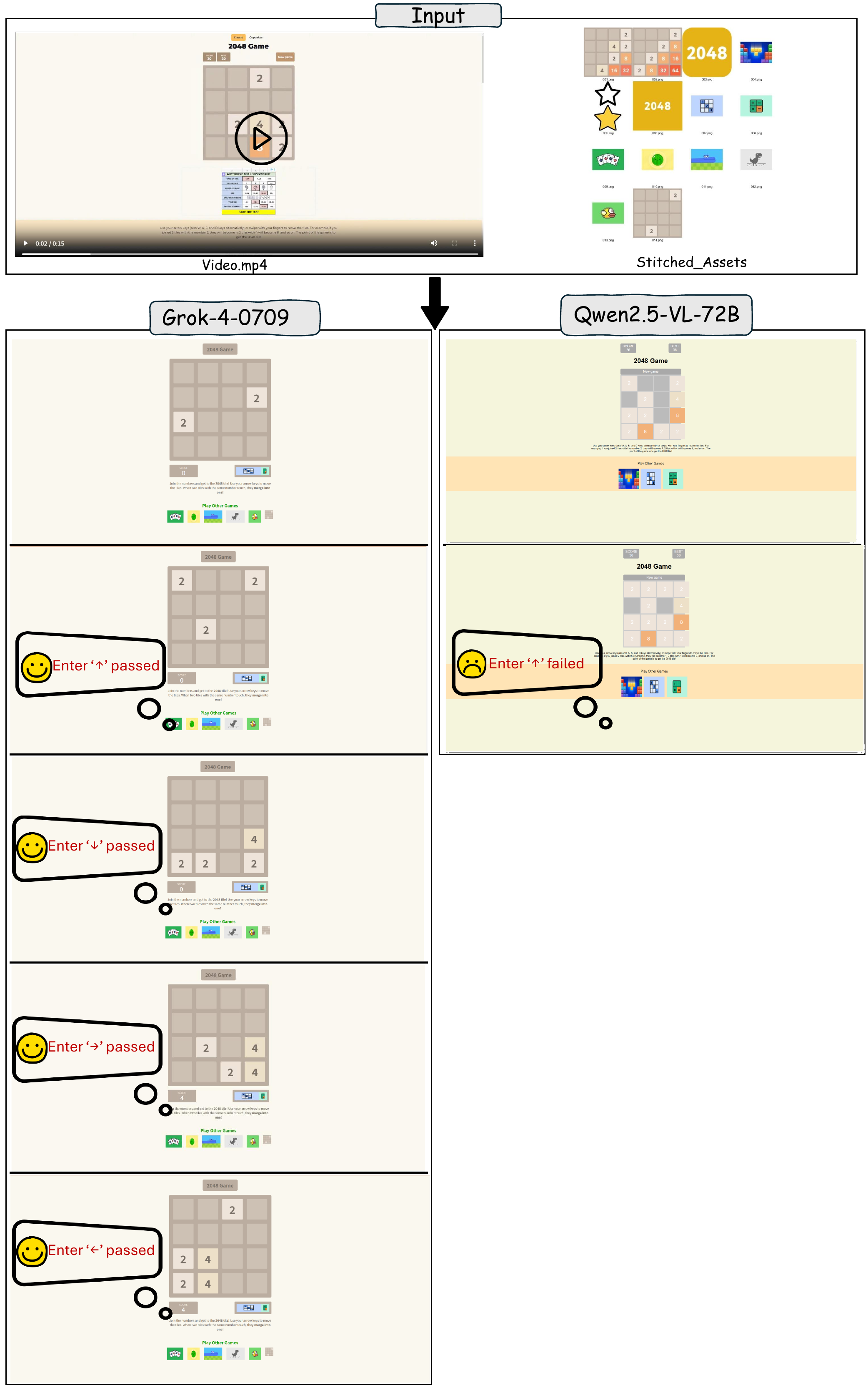}
    \caption{
        \textbf{Case 2: Algorithmic Game Logic Reconstruction.} This task challenges models to reconstruct the rules of the simple browser game 2048. Classified as \textbf{[L4, V2, Gaming]}, the primary difficulty lies in algorithmic correctness, not visual complexity. 
    }
    \label{fig:case2}
\end{figure}

As illustrated in Figure \ref{fig:case2}, the left side is the result of Grok-4, which can successfully reproduce the 2048 game logic from the input video. However, Qwen2.5-VL-72B failed to merge the corresponding blocks after inputting '↑'. This type of task requires a high level of logical reasoning ability from the model and is a significant challenge.

\paragraph{Case 3 Analysis: Long-Context Fidelity and Fine-Grained Visual Detail.}
This case study, classified as [L1, V3, E-commerce], is designed to stress-test a model's visual fidelity on multiple fronts. While the interaction is simple (L1, passive scrolling), the task's difficulty lies in three distinct challenges: (1) maintaining structural integrity across a long page, (2) correctly handling diverse media assets, including an embedded video, and (3) achieving fine-grained visual accuracy, particularly with small, repetitive elements like icons. This multi-faceted task evaluates not just broad layout reconstruction but also the model's attention to detail and its ability to precisely match visual elements to the provided stitched assets.

\begin{figure}[h]
    \centering
    \includegraphics[width=0.99\linewidth]{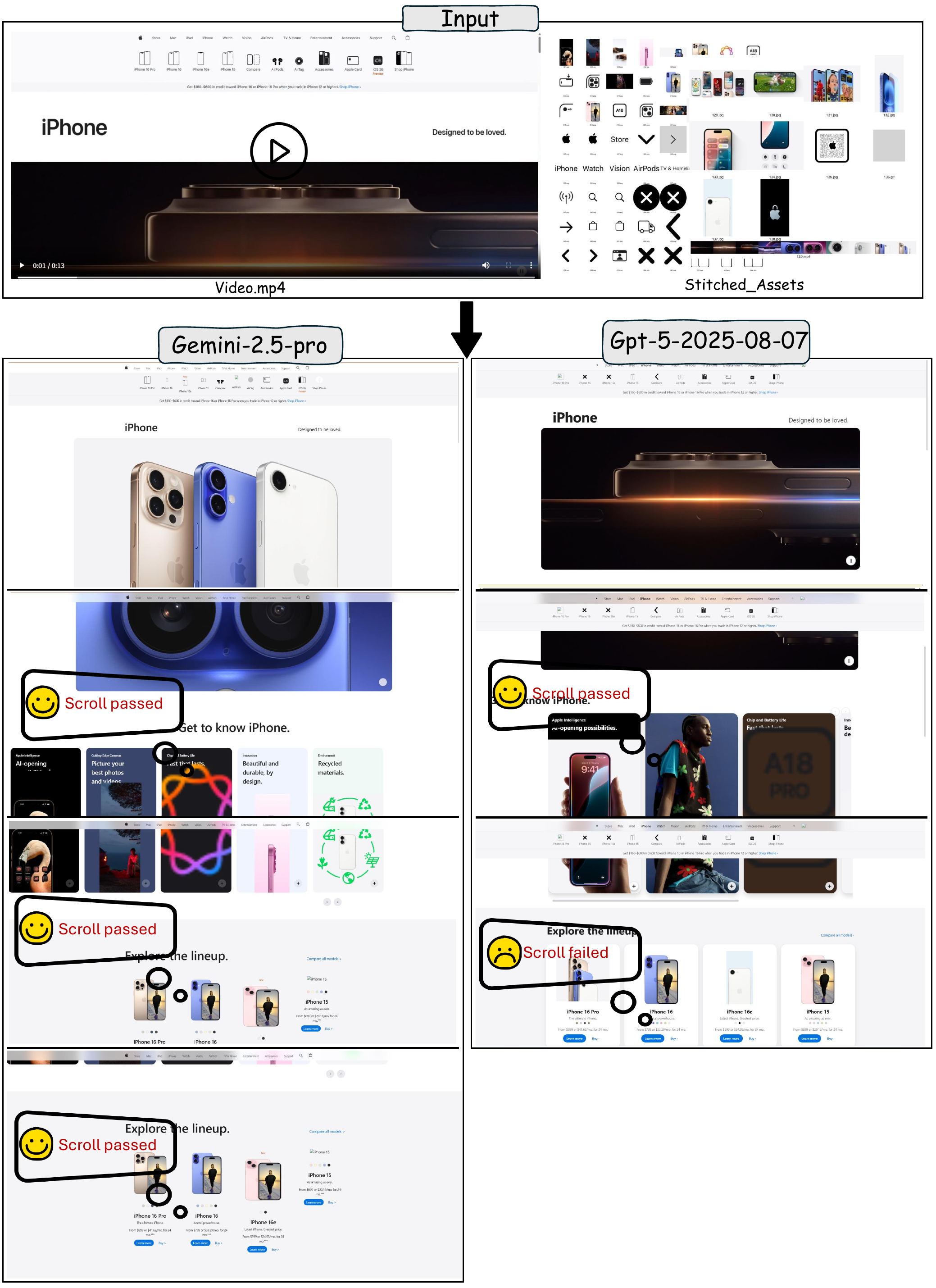}
    \caption{
        \textbf{Case 3: Full-Page Reconstruction with Long Scrolling.} This task focuses on a fundamental capability: reconstructing a webpage that extends far beyond the initial viewport. Classified as \textbf{[L1, V3, E-commerce]}, it tests the model's ability to handle static content at scale. 
    }
    \label{fig:case3}
\end{figure}

As illustrated in Figure \ref{fig:case3}, Gemini-2.5-pro and GPT-5 can both restore relatively complete long pages based on videos, but they do not handle the details of the web pages well, including the corresponding icons and matching product images. A successful reconstruction would require both holistic understanding of the page structure and meticulous attention to its smallest components.

\paragraph{Case 4 Analysis: Time-Based State Management in a Mobile Viewport.}
This case study, classified as [L3, V1, Productivity \& Tools], is designed to evaluate a model's ability to handle time-driven state changes, presented within the constraints of a mobile screen resolution. The task is to reconstruct a functional Pomodoro timer. While the visual complexity is low (V1), the mobile viewport requires the model to generate a layout that is responsive or appropriately scaled for a narrow screen. The primary challenge, however, resides in the L3 interaction complexity: the model must implement a state logic governed by both user clicks (e.g., 'start', 'pause') and asynchronous, time-based events (the countdown reaching zero).

\begin{figure}[h]
    \centering
    \includegraphics[width=0.99\linewidth]{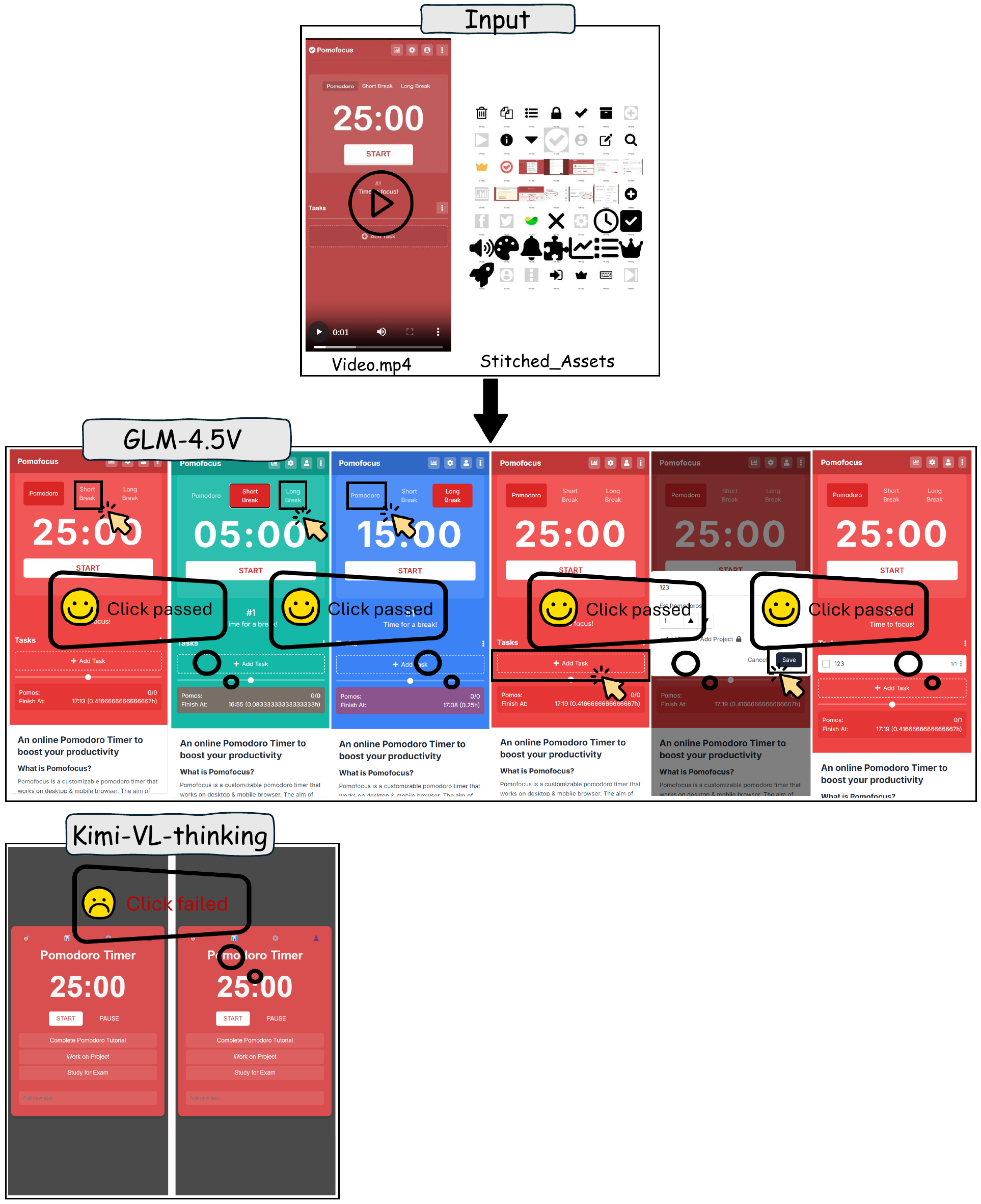}
    \caption{
        \textbf{Case 4: Pomodoro Timer Logic within a Mobile Viewport.} This task requires reconstructing a Pomodoro timer rendered at a mobile resolution. Classified as \textbf{[L3, V1, Productivity \& Tools]}, the core challenge is not the simple layout but the implementation of time-based state transitions (start, pause, reset).
    }
    \label{fig:case4}
\end{figure}

As illustrated in Figure \ref{fig:case4}, GLM-4.5V can successfully implement the interactive operations and logic in the video, but Kimi-VL-thinking is unable to perform subsequent operations because the elements that need to be clicked in the first step are missing in the initial state.

\section{USE OF LARGE LANGUAGE MODELS}
\label{sec:app_llm_use}
We utilized a Large Language Model to assist with grammar correction and language refinement in this paper.



\end{document}